\documentclass[runningheads]{llncs}
\usepackage[T1]{fontenc}
\usepackage{graphicx}
\usepackage{booktabs}
\usepackage{dirtytalk}
\usepackage{url}
\usepackage{amsmath}
\usepackage{hyperref}
\usepackage{orcidlink}
\usepackage{cite}
\providecommand{\orcidID}{}
\renewcommand{\orcidID}[1]{\,\orcidlink{#1}}

\begin{document}
\title{ANTShapes Benchmarking Datasets for Event-Based Neuromorphic Object Classification}
%
%
\author{M.~Middleton*\inst{1}\orcidID{0000-0002-0862-7524} \and
H.~Kayan\inst{2}\orcidID{0000-0002-4878-2326} \and
B.~Sen~Bhattacharya\inst{3}\orcidID{0000-0001-5079-0619} \and
T.~Ali\inst{4}\orcidID{0000-0002-1576-6645} \and
E.~Baikas\inst{5}\orcidID{0009-0007-0376-7636} \and
M.~Vousden\inst{5}\orcidID{0000-0002-6552-5831} \and
C.~Perera\inst{2}\orcidID{0000-0002-0190-3346} \and
O.~Rhodes\inst{3}\orcidID{0000-0003-1728-2828} \and
E.~Gheorghiu\inst{4}\orcidID{0000-0002-9459-1969} \and
M.~A.~Trefzer\inst{1}\orcidID{0000-0002-6196-6832}}
\authorrunning{M. Middleton et al.}
%
\institute{University of York, UK \\ \email{\{michael.middleton\}, \{martin.trefzer\}@york.ac.uk} \and
Cardiff University, UK \\ \email{\{kayanh\}, \{pererac\}@cardiff.ac.uk} \and
University of Manchester, UK \\ \email{\{basab\}, \{oliver.rhodes\}@manchester.ac.uk} \and
University of Stirling, UK \\ \email{\{t.r.ali\}, \{elena.gheorghiu\}@stir.ac.uk} \and
University of Southampton, UK \\ \email{\{e.baikas\}, \{m.vousden\}@soton.ac.uk}
}
\maketitle              
\begin{abstract}

Object classification in event-based computer vision is a task that is attracting considerable research attention.
Event-based object classification is a fundamental task in the fields of security and applied computer vision, which typically use synchronous frame-based cameras and computing pipelines for operation.
This approach has several practical flaws.
The size, weight and power consumption of the device could prohibit deployment at the extreme edge or in covert sensing environments.
Besides this, there are security concerns inherent in cloud-based or other \say{off-device} computation approaches due to the requirement of sending and receiving potentially sensitive data.
Furthermore, this transmission of data introduces latency and requires consistent connectivity to the cloud infrastructure to function.

The use of Spiking Neural Networks (SNNs) hosted on neuromorphic devices attempts to solve several issues present in this conventional approach.
Research into event-based object classification methods are hindered by the lack of high-quality vision datasets to use.
To this end, the ANTShapes simulation tool has been previously proposed to create and label event-based vision datasets.
In this paper, four novel datasets of varying \say{difficulties} are created using the tool and are benchmarked against existing spiking datasets commonly used for event-based vision research (N-MNIST, CIFAR10-DVS, DVSGesture and POKER-DVS).
Classification is performed using a convolutional SNN.
This work simultaneously provides four datasets with rich details for future experiments to use and validates the output of the ANTShapes dataset simulation tool as being suitable for its purpose.

\end{abstract}

\section{Introduction}

Spiking Neural Networks (SNNs) promise a low-power alternative to traditional Artificial Neural Networks (ANNs) by their asynchronous approach to performing computations~\cite{maass1997networks, roy2019spikebased}.
As a bio-inspired approach, processing units (or neurons) within a neuromorphic computing device (such as Intel's Loihi chips~\cite{davies2018loihi}, SpiNNaker~\cite{painkras2013spinnaker} and BrainScaleS~\cite{schemmel2020brainscales}) only consume power when performing computations. 

This event-driven approach to computation is a fundamentally different approach to computation on Von Neumann architectures using conventional clock-driven processors, where power is dissipated continuously regardless of whether meaningful computation is occurring~\cite{indiveri2015memoryandinformationprocessing}.
Inference from SNNs hosted on neuromorphic hardware can be performed at microsecond time scales and at a fraction of the power budget required by conventional deep learning hardware~\cite{davies2018loihi, taunyazov2020eventdrivenvisualtactilesensinglearning, cramer2022surrogategradients}.
As a result, SNNs are particularly well-suited for deployment at the extreme edge or in other instances where size, weight and power of the device must be carefully managed, or when low latency operation is a high priority consideration~\cite{kudithipudi2025neuromorphiccomputingatscale, cheng2026braininspiredaiedgeintelligence}.

The potential energy efficiency and inference speeds of neuromorphic computing devices make them especially attractive for computer vision tasks where inference must often be performed continuously and in real time.
Object classification represents one of the most fundamental such tasks, forming the basis of broader applications including scene understanding \cite{zhang2024accurateefficienteventbasedsemantic, zhang2023energyefficientspikingsegmenter}, autonomous navigation for vehicles and robots \cite{viale2022lanesnns, milde2017obstacleavoidance} and anomaly detection in security \cite{qian2025ucfcrimedvs} and monitoring \cite{shariff2024spikingdd, zhu2025ultrafastdynamicdefect} contexts. 
In the context of anomaly detection, an accurate object classifier serves as a prior for defining normality.
Deviations from expectations in the classes of objects populating a scene or in the temporal dynamics (alternatively, the \say{event signatures}) expressed by objects can then be flagged as anomalous events. 
This has significant practical implications across industrial quality control, surveillance and safety-critical monitoring, where low power consumption and latency are equally important constraints~\cite{haessig2019spiking, zhu2025ultrafastdynamicdefect}.

Currently, there is a shortage of suitable benchmark datasets for object classification using neuromorphic devices~\cite{gallego2022eventbasedvision, middleton2026eventbasedvisionattheedge}.
Existing neuromorphic datasets present a number of limitations that restrict their utility for evaluating object recognition pipelines.
These limitations include the absence of true spatiotemporal features encoded into data from naturally evolving scene dynamics and the lack of purpose-built neuromorphic vision datasets for the classification of 3D objects. 
There remains a clear need for a purpose-built, noise-characterised, temporally rich neuromorphic datasets that directly addresses the object classification task.

\section{Background}
\label{ss:background}

To understand the needs of a proposed neuromorphic vision dataset for object classification, four datasets which are frequently used in literature are examined here.
The scope is restricted to the N-MNIST, CIFAR10-DVS, POKER-DVS and DVSGesture datasets which are often studied in literature for event-based classification tasks~\cite{gallego2022eventbasedvision}.

\subsection{N-MNIST}
\label{ss:nmnist}

\begin{figure}[h!]
    \centering
    \includegraphics[width=0.7\linewidth]{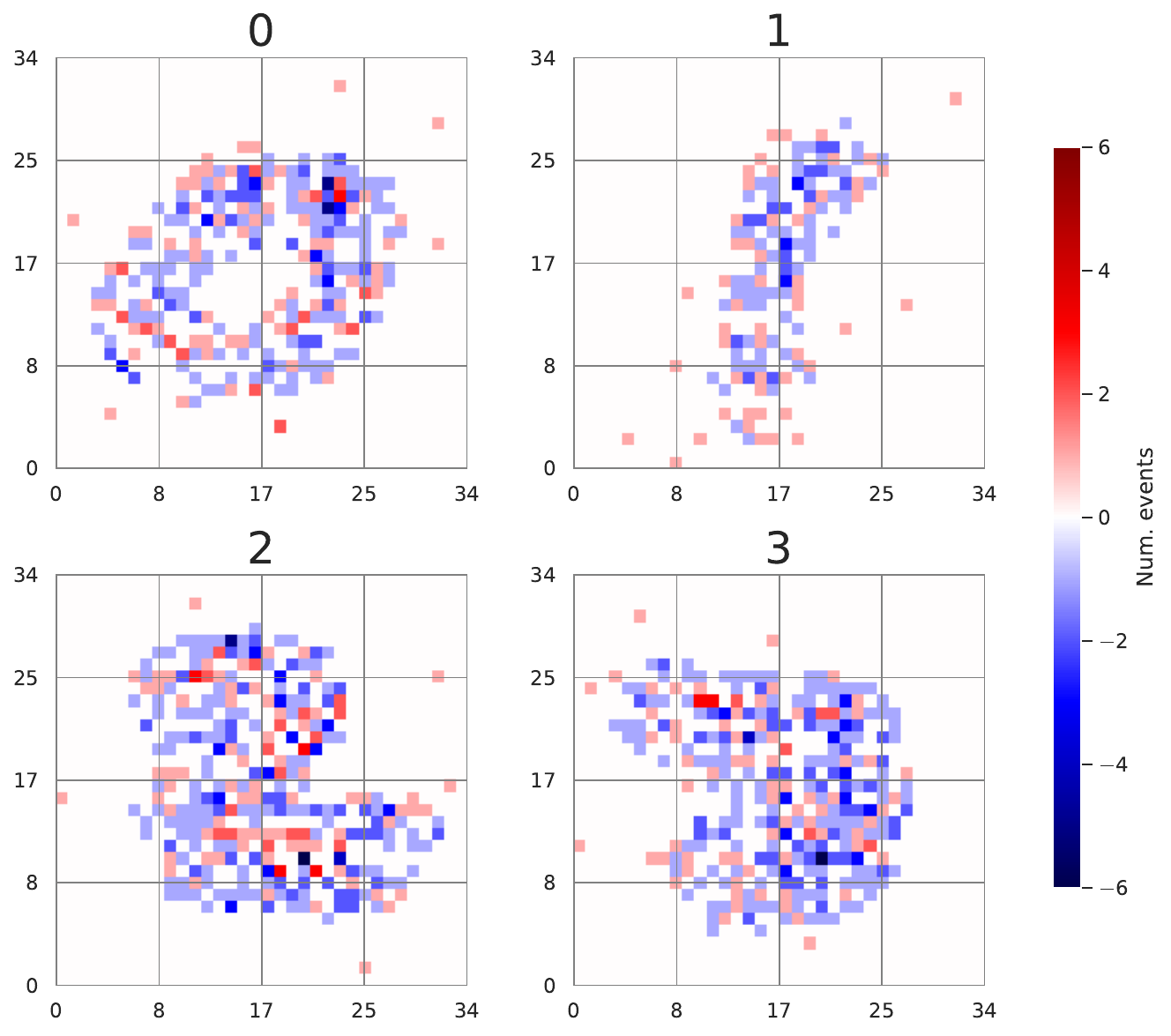}
    \caption{Examples of the 0, 1, 2 and 3 classes taken from the N-MNIST dataset.  A denoising process (explained later in Section~\ref{ss:dataset_preparation} is applied before events are accumulated per-pixel for display.  Pixels with more "on" than "off" events are coloured red whilst pixels of the opposite case are coloured blue.}
    \label{fig:nmnist_example}
\end{figure}

The Neuromorphic MNIST (N-MNIST) dataset is an event-based representation of the MNIST handwritten digits dataset. 
It was produced by mounting an Asynchronous Time-based Image
Sensor (ATIS) event camera on a pan-tilt actuated platform and moving it in front of an LCD monitor displaying static MNIST digit images~\cite{orchard2015converting}. 
The camera performed three pre-defined smooth movements, or saccades, roughly emulating the saccadic movements of the eye.
Each saccade lasts approximately 105 ms and takes the path of an isosceles triangle.
Examples of digits 0 to 3 from the N-MNIST dataset are visualised in Figure~\ref{fig:nmnist_example}.

The relative motion between the sensor and the displayed image induces local changes in pixel intensity, which the DVS encodes asynchronously as a stream of events in the Address-Event Representation (AER) format. 
Each event is a tuple $(x, y, t, p)$, where $x$ and $y$ denote the pixel address within the $34 \times 34$ sensor array, $t$ is a microsecond-resolution
timestamp and $p \in \{+1, -1\}$ is the polarity, indicating whether local log-intensity increased (\say{on} event) or decreased (\say{off} event). 
On average, each recording contains approximately 4,172 events and spans 360 ms in total. 
The dataset mirrors the structure of the original spatial MNIST dataset, consisting of 60,000 training and 10,000 test samples.
Accordingly, there are 10 classes represented in the N-MNIST dataset for the digits 0 to 9.

N-MNIST has been used for several studies into neuromorphic computer vision as a proxy for object classification~\cite{fang2021plif, eshraghian2023snntorch}.
The reasoning for this is clear; the N-MNIST digits contain distinct features in space for each of the 10 digit classes.
Therefore, parallels can be drawn between digit and object recognition tasks as both require the separation of features present in the visual data.

However, it has also been noted that N-MNIST contains little meaningful temporal information as the scene content remains static over time, with events generated from the motion of the dynamic vision sensor~\cite{iyer2021nmnist}.
Fundamentally, a digit from the dataset is a two-dimensional image which exists as surface detail on the object it is written on.
Although digits can be manipulated to produce variance in spatial detail (such as rotating, flipping or skewing digits, or animating a motion path for the digit to follow), the spatial data produced cannot be as rich as the potential data represented in a 3D object recognition dataset where objects may be viewed from any number of angles and features between classes can be similar or dissimilar based on the viewing angle.
The distinction between 2D image and 3D object classification can be formalised on this basis.
Whilst the fundamentals of detail and object classification are broadly similar, the inability to represent variance in viewing angles for 2D datasets such as N-MNIST demonstrates that they are not ideal substitutions for object classification tasks.

\subsection{CIFAR10-DVS}

\begin{figure}[h!]
    \centering
    \includegraphics[width=0.7\linewidth]{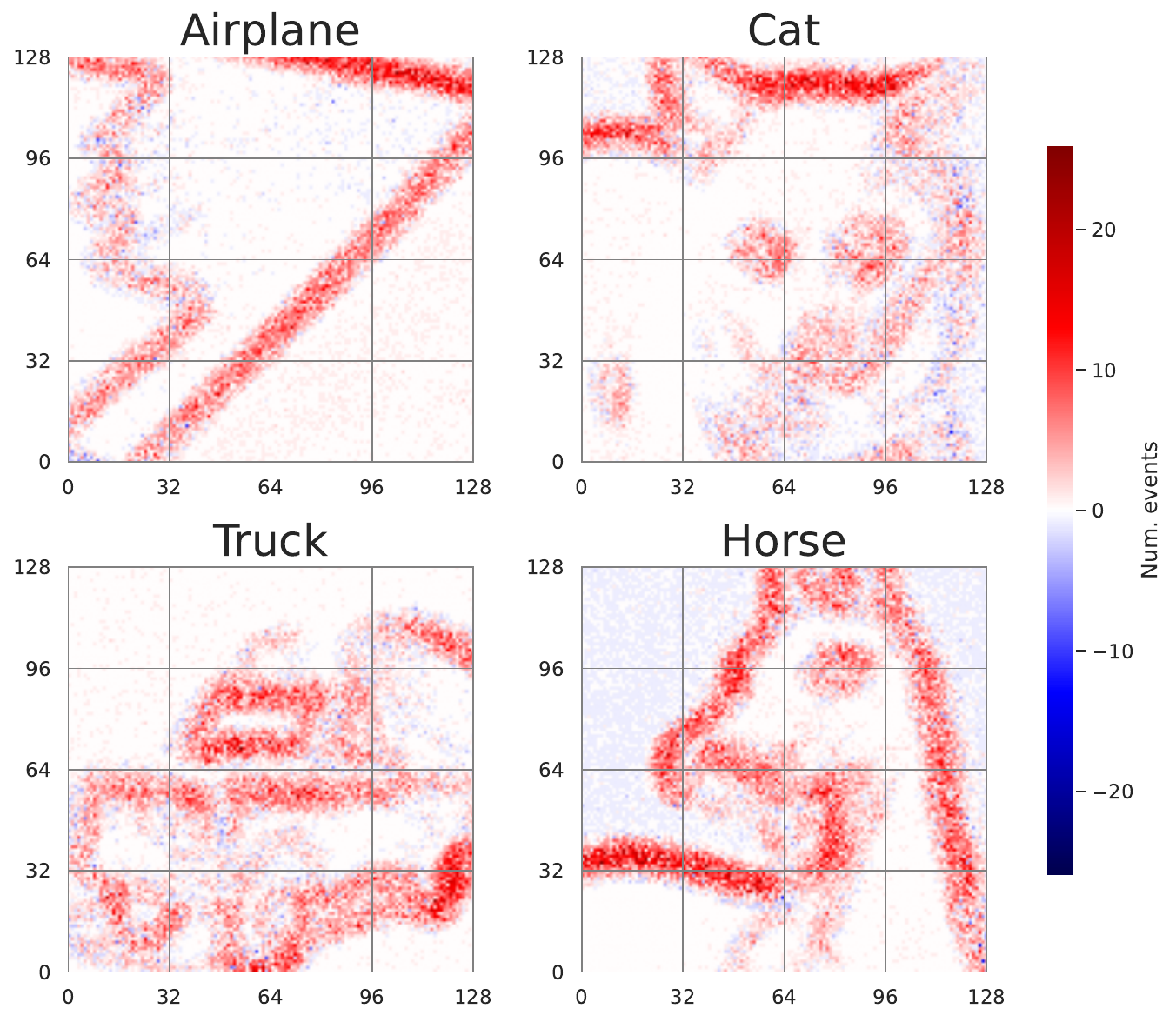}
    \caption{Examples of the "airplane", "cat", "truck" and "horse" classes taken from the CIFAR10-DVS dataset.}
    \label{fig:cifar10dvs_example}
\end{figure}

CIFAR10-DVS~\cite{li2017cifar10dvs} is an event-stream counterpart of the CIFAR-10 image classification dataset, designed to offer a benchmark of intermediate difficulty for neuromorphic vision research. 
The dataset was produced by displaying 1,000 images per class (10,000 in total) on an LCD monitor and moving them in a \say{repeated closed-loop smooth} (RCLS) pattern within the field of view of a DVS camera with a sensor resolution of $128\times128$ pixels. 
This is also to emulate saccadic eye movements, although the approach taken here moves the on-screen image itself rather than the DVS.
Examples from the CIFAR10-DVS dataset are visualised in Figure \ref{fig:cifar10dvs_example}.

Prior to display, the original $32\times32$ CIFAR-10 images were upscaled to $512\times512$ via bicubic interpolation to ensure sufficient contrast when
rendered on the monitor. 
The resulting event streams contain richer spatiotemporal structure than their MNIST-derived counterparts, reflecting the more complex textures and object classes described in the CIFAR10 dataset (aeroplanes, automobiles, birds, cats, deer, dogs, frogs, horses, ships and trucks).

Like N-MNIST, the CIFAR10-DVS dataset can be viewed as a detail classification dataset rather than true object classification.
Although the classes represented in CIFAR10-DVS are real objects that can exist independently in the real world (unlike a hand-drawn digit), the method of producing the dataset is similar to N-MNIST; no scene evolution is encoded into the source CIFAR10 dataset so the data represented is spatial in nature, rather than spatiotemporal.
This limits the application of CIFAR10-DVS towards object classification in neuromorphic contexts in a similar manner as explored in Section \ref{ss:nmnist}.
Furthermore, as shown in Figure \ref{fig:cifar10dvs_example}, features are dominated by \say{on} polarity events.
This may pose issues during the training of an SNN if both polarity channels are treated with equal importance due to the relative sparsity of spikes in the \say{off} channel.

\subsection{POKER-DVS}

\begin{figure}[h!]
    \centering
    \includegraphics[width=0.7\linewidth]{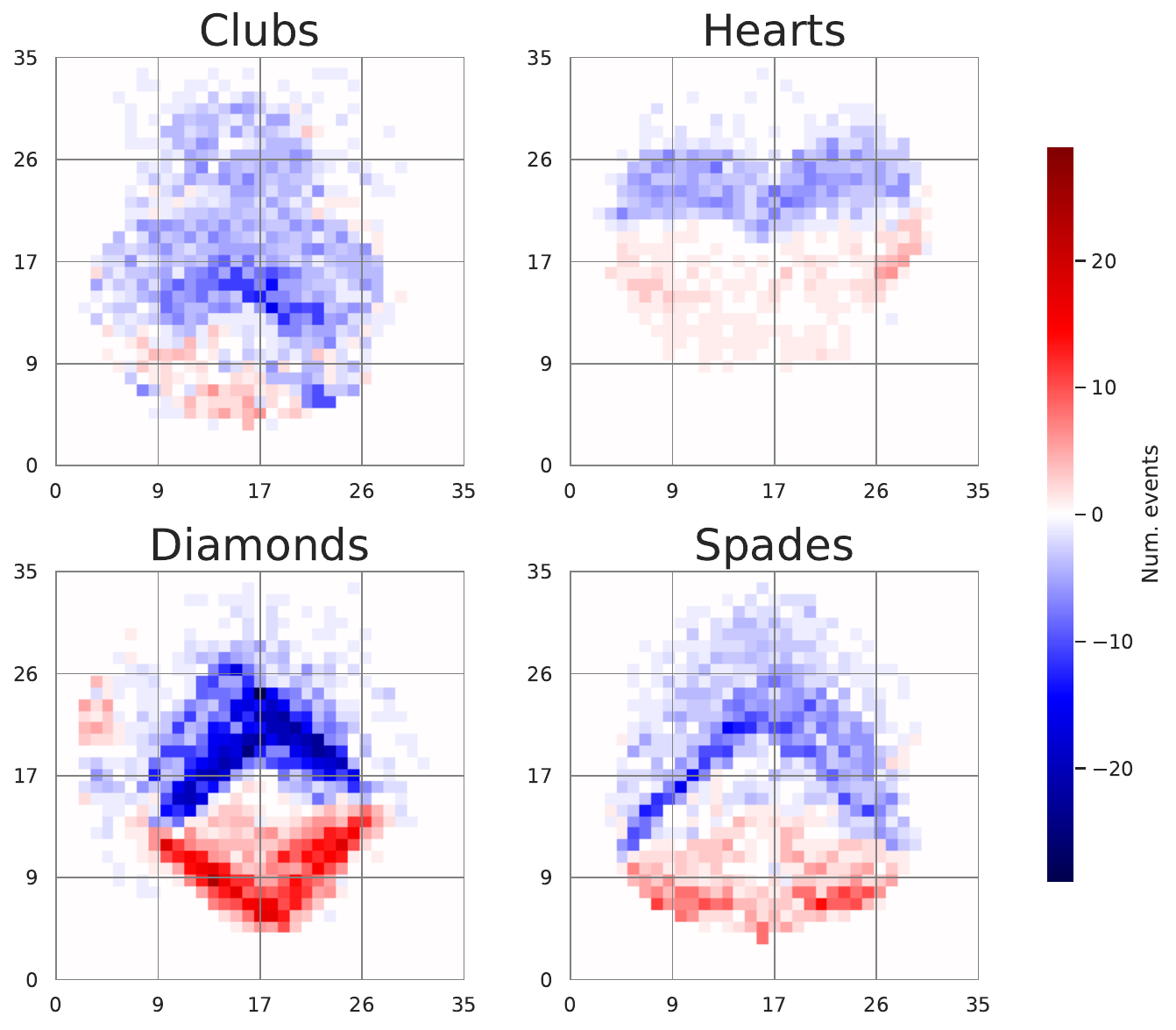}
    \caption{Examples of the four classes of poker card symbols taken from the POKER-DVS dataset.  The resolution of $35\times35$ pixels shown here is used in practice for POKER-DVS rather than the published value of $32\times32$.  Some Python implementations of the dataset, such as the Tonic package implementation, use a resolution of $35\times35$ pixels.}
    \label{fig:pokerdvs_example}
\end{figure}

The POKER-DVS dataset was introduced as a benchmark for high-speed event-driven object recognition~\cite{serrano2015poker}.
It was recorded by rapidly browsing a custom-made poker card deck in front of a DVS at close range.
Each card passes through the sensor's $128\times128$ pixel field of view in approximately 20-30 ms, producing dense streams of events at rates approaching 8 mega-events per second.
Individual playing card suits (clubs, diamonds, hearts and spades) were subsequently tracked and isolated offline to yield 131 labelled recordings, each nominally cropped to a $32\times32$ pixel window centred on the card suit.
Similarly to the N-MNIST dataset, examples in the POKER-DVS dataset are represented in AER format.
Figure \ref{fig:pokerdvs_example} shows examples from each of the four classes of suits included in the dataset.

POKER-DVS is similar to the other datasets explored so far, albeit with a smaller number of classes compared to the 10 present in N-MNIST and CIFAR10-DVS.  
The poker symbols extracted from the DVS data represent surface detail on a 3D object and may not be suitable as a proxy for object classification.
However, POKER-DVS differentiates itself by being sourced from real-world data, as opposed to converting a purely spatial dataset using a DVS.
Therefore, POKER-DVS contains true spatiotemporal features in its data.

\subsection{DVSGesture}

\begin{figure}[h!]
    \centering
    \includegraphics[width=0.7\linewidth]{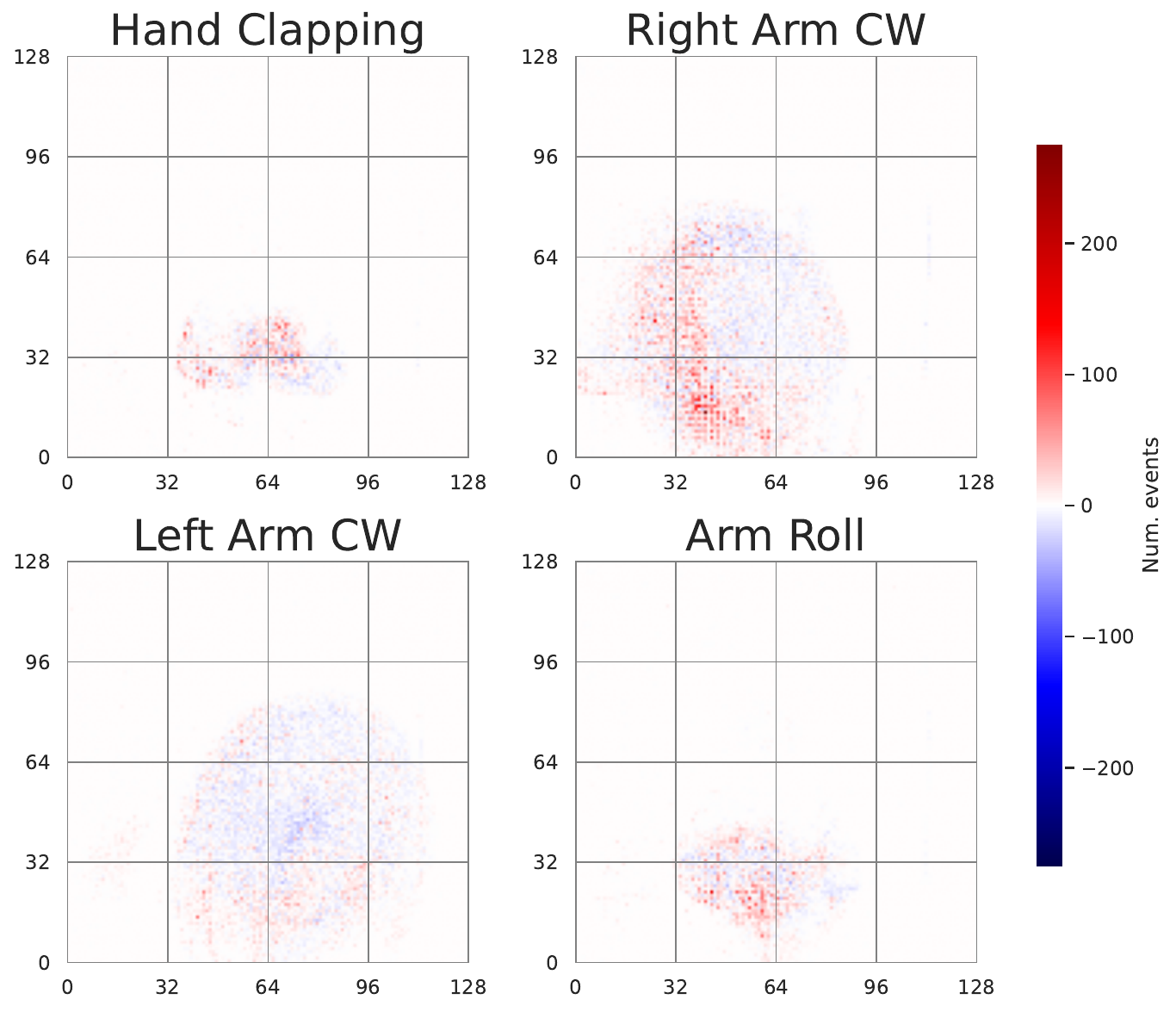}
    \caption{Examples of the "hand clapping", "right arm CW", "left arm CW" and "arm roll" classes taken from the DVSGesture dataset.  The right and left arm gestures represent moving the relevant arm in a circular motion clockwise.}
    \label{fig:dvsgesture_example}
\end{figure}

The DVSGesture dataset~\cite{amir2017low} is a natively event-based action recognition dataset, distinct from the converted datasets described above in that its events arise from real dynamic scenes rather than static images
displayed on a monitor. 
It was recorded using the iniVation DVS128 camera at a $128\times128$ pixel resolution, capturing 29 subjects each performing 11 hand and arm gestures in a single sequential trial under three different illumination conditions (natural
light, fluorescent light, and LED light), yielding 1,342 gesture instances in total.
The 11 gesture classes include hand waving with both arms, large clockwise and
counter-clockwise arm rotations, forward and backward forearm rolling, air guitar, air drums, and an open-ended \say{other} class for miscellaneous gestures which were chosen at the discretion of the performers.
Gestures from the dataset are visualised in Figure \ref{fig:dvsgesture_example}.

Like POKER-DVS, DVSGesture is sourced from real-world data using a DVS camera so the samples within contain true spatiotemporal events.
However, the actions expressed by individuals are not a perfect analog to the events produced by different object classes in action.
Strictly, the human shown in DVSGesture scenes is the \say{object} in motion and the gestures expressed represent different classes of action the object can experience.
An equivalent scenario using a primitive object (such as a cube) could contain different patterns of rotation, translation and so on.
The object itself does not change, only its actions.
This illustrates the difference between action and object classification and suggests the DVSGesture dataset is not a suitable proxy to be used for evaluating object classification tasks using SNNs.

\subsection{Summary of Existing Datasets}

An overview of the four datasets studied here is presented in Table \ref{tab:dataset_summary}.  
It can clearly be seen that none target object classification specifically and of the two which contain natural features from evolving scenes, neither are specifically targeted towards object classification.

\begin{table}[h!]
\resizebox{\columnwidth}{!}{%
\begin{tabular}{|l|l|l|l|l|l|}
\hline
\textbf{Dataset} & \textbf{Purpose}                                                           & \textbf{\begin{tabular}[c]{@{}l@{}}Spatial\\ Resolution\end{tabular}} & \textbf{Num. Classes}                                         & \textbf{\begin{tabular}[c]{@{}l@{}}Type of\\ Features\end{tabular}} & \textbf{\begin{tabular}[c]{@{}l@{}}Contains\\ DVS Noise?\end{tabular}} \\ \hline
N-MNIST          & \begin{tabular}[c]{@{}l@{}}Handwritten\\ digit classification\end{tabular} & $34 \times 34$                                                        & 10                                                            & \begin{tabular}[c]{@{}l@{}}Saccadic\\ eye mimic\end{tabular}        & Yes                                                                    \\ \hline
CIFAR10-DVS      & Image classification                                                      & $128 \times 128$                                                      & 10                                                            & \begin{tabular}[c]{@{}l@{}}Saccadic\\ eye mimic\end{tabular}        & Yes                                                                    \\ \hline
POKER-DVS        & \begin{tabular}[c]{@{}l@{}}Object detail\\ classification\end{tabular}     & $32 \times 32$                                                        & 4                                                             & \begin{tabular}[c]{@{}l@{}}Natural\\ scene\end{tabular}             & Yes                                                                    \\ \hline
DVSGesture       & \begin{tabular}[c]{@{}l@{}}Object detail\\ classification\end{tabular}     & $128 \times 128$                                                      & \begin{tabular}[c]{@{}l@{}}10 \\ + 1 \say{other}\end{tabular} & \begin{tabular}[c]{@{}l@{}}Natural\\ scene\end{tabular}             & Yes                                                                    \\ \hline
\end{tabular}%
}
\hfill
\caption{Overview of the four existing datasets discussed here.  Only one dataset, CIFAR10-DVS, is specifically targeted towards the classification of an entire object rather than just details expressed on or by an object (such as suit printed on a card or gesture expressed by a person's body).}
\label{tab:dataset_summary}
\end{table}

The N-MNIST and CIFAR10-DVS datasets both produce spatiotemporal events from spatial image data by mimicking the saccadic movements produced by the eye.
This is a valid method of translating spatial data into spatiotemporal event patterns, although the data contained in the dataset remains strictly spatial in character.
Therefore, these datasets contain no meaningful temporal structure within samples, as the apparent motion is artificially induced by camera movement rather than arising from natural scene dynamics~\cite{iyer2021nmnist}.

Furthermore, the saccadic patterns produced by the movement of the DVS for N-MNIST or the translation of the image on the monitor itself for CIFAR10-DVS follow fixed, programmatic trajectories. 
Consequently, the temporal structure of the resulting spike trains is an artefact of the recording motion rather than a property of the object class itself and all samples share the same broad event signature dynamics irrespective of their content. 

All neuromorphic datasets produced with DVS cameras are also affected by inherent sensor noise. 
DVS pixels fire asynchronously in response to changes in log-luminance, causing thermal noise and junction leakage to produce spurious spike events that cannot be attributed to an actual change in scene luminance~\cite{brandli2014dvs}. 
This noise is present throughout standard benchmarking datasets explored in this paper.
Whilst event-based noise reduction processes can filter some of this spurious noise away, ideally any noise present in a neuromorphic object classification dataset should be minimised as much as possible and tightly controlled wherever else.
This would prevent any features relevant to the data in the scene from being removed in the process and provide an idealised scenario to study in the process of developing neuromorphic vision systems.

Finally, datasets such as POKER-DVS~\cite{serrano2015poker} and DVSGesture, while widely cited in the neuromorphic computing literature, are not strictly object classification benchmarks.
For example, POKER-DVS is designed to evaluate the detection of suit symbols on playing cards under constrained conditions.
Whilst visual classification can be performed on the four classes of suits, the symbols are two-dimensional printed features on the surfaces of moving playing cards.
Alternatively, DVSGesture targets the classification of hand and arm gestures.
The dominant features of samples from DVSGesture arise from the movement of limbs attached to a body positioned in frame, rather than from the identity of a fully discrete object class. 
Therefore, gesture classification is not an ideal proxy for object classification.

From this discussion, it is evident that there is a necessity for an object classification dataset for the benchmarking of SNN computer vision pipelines.
The criteria for such a dataset is as follows:

\begin{itemize}
    \item \textbf{True spatiotemporal data:} the dataset must contain natural features that arise as a description of the scene as it changes over time.
    \item \textbf{DVS-style data representation:} the dataset must be fundamentally comparable to data captured from real-world scenes using a DVS.
    \item \textbf{Free from spurious noise:} the dataset must be free from stochastic noise, as produced by dynamic vision sensors, to provide clean data for study.
\end{itemize}

The simulated datasets proposed in the following section aim to address these identified needs.

\section{ANTShapes Dataset Generation}

To fulfil the desired criteria specified above, it was decided that the benchmarking datasets proposed here would be simulated computationally rather than captured using a DVS.
This is to eliminate the stochastic noise captured by the DVS and to allow for tight control over the features represented in the dataset by means of parameterisation.
The ANTShapes (Anomalous Neuromorphic Tool for Shapes) simulation software is used for this purpose as it can represent 12 distinct object classes with rich spatiotemporal dynamics for object classification~\cite{middleton2026antshapes}.
The classes of shapes included in the dataset are as follows:

\begin{itemize}
    \item \textbf{Flat-faced objects:} cuboids, icospheroids, pyramids, \say{Tetris}-style L and T-blocks.
    \item \textbf{Smooth-faced objects:} spheroids, cylinders, cones, toroids, capsules.
    \item \textbf{Complex meshes:} Utah teapot and Suzanne monkey head reference meshes.
\end{itemize}

It is noted that datasets produced by the ANTShapes software has not been validated by experimental means~\cite{middleton2026antshapes}.
The experiments presented here also aim to demonstrate that the software is fit for its designed purpose of simulating scenes populated by simple 3D objects for event-based computer vision tasks.
Principal Component Analysis (PCA) of the event data produced for each labelled sample will also indicate the degree of separation between class features.
PCA is applied using the \texttt{sklearn} package in Python and the two highest principal components of samples are clustered for visualisation; these plots are shown in Section \ref{ss:apx}.

Simulated data is exported in Address Event Representation (AER) format, similarly to DVS sensors, where changes in scene luminance over time are represented as packets of data containing the X and Y screen coordinate, timestamp and \say{on} or \say{off} polarity. 

Four benchmarking datasets are produced in total.
The first is labelled as the \say{standard} dataset, which is intended to be the baseline ANTShapes dataset.
Three additional datasets are derived from this standard dataset, with each presenting some additional challenge by introducing some additional behaviour to simulated objects.
An overview of each ANTShapes dataset is as follows:

\begin{itemize}
    \item \textbf{Standard:} 1 of 12 discrete object classes is presented in an empty scene with some variation in its $(x, y)$ position.  Each object begins with some random rotation around the $(x, y, z)$ axes.  Objects continue to rotate around all three axes over time as defined by a constant unit vector, producing natural features from changes in scene log-luminance.
    \item \textbf{Shape Translation:} The 3D objects also translate across the scene with some random $(x, y)$ velocity which is independent for each simulated example.  When approaching the edge of the scene, the objects \say{bounce} back into the view of the virtual DVS so objects never leave the simulated area. 
    \item \textbf{Shape Distortion:} Objects are scaled around the $(x, y, z)$ axis independently by some random 3D vector.  This distorts the proportions of each rendered object, squashing and stretching along each axis and altering the spatiotemporal patterns produced by the object's rotation over time.
    \item \textbf{Random Rotation:} Rather than rotating each shape around the $(x, y, z)$ axes by a constant unit vector, a random rotation vector is used for each rendered example.  This alters the spatiotemporal patterns produced between examples by introducing variation to the streams of events generated by the object's motion.
\end{itemize}

Each ANTShapes dataset contains 3,000 examples (or $\approx 250$ examples per object class) and is rendered at a temporal resolution of 1 ms per event.
All rendered examples are 836 ms long and a shape rotates around its axes once every 209 ms, assuming a unit rotation vector of $(1, 1, 1)$.
Lighting between rendered scenes is constant and rendered as a neutral-toned global light source approximating outdoor conditions.

\subsection{Standard ANTShapes Benchmarking Dataset}

The software parameters used to simulate this dataset are provided in Table \ref{tab:antshapes_params}.
Parameters not described in this table are left at their default values.

\begin{table}[h!]
\centering
\resizebox{0.8\columnwidth}{!}{%
\begin{tabular}{|l|l|l|}
\hline
\textbf{Parameter}                     & \textbf{Setting} & \textbf{Description}                                                                                                                                                                                                                                   \\ \hline
Initial Rotation $\sigma$, $(x, y, z)$ & 180              & \begin{tabular}[c]{@{}l@{}}Randomises the initial rotation\\ position of the object around\\ the $(x, y, z)$ axes.\end{tabular}                                                                                                                        \\ \hline
Initial Position $\sigma$ $(x, y)$     & 0.15             & \begin{tabular}[c]{@{}l@{}}Displaces the shape from the\\ center of the scene on the \\ horizontal and vertical axes.  \\ This provides variance in the \\ dataset by randomising object \\ positions within scenes over \\ each example.\end{tabular} \\ \hline
Scale $\mu$ $(x, y, z)$                & 4                & \begin{tabular}[c]{@{}l@{}}A constant scaling value is \\ applied to all shapes to \\ enlarge them in the scene.  \\ Scaling is applied uniformly \\ across the $(x, y, z)$ axes.\end{tabular}                                                         \\ \hline
Surface Noise $\mu$                    & 0.2              & \begin{tabular}[c]{@{}l@{}}Applies normal-mapped noise\\  to the surfaces of objects.  \\ This applies visual texture \\ to shapes which is dependent \\ on the incident angle of the \\ scene light source.\end{tabular}                              \\ \hline
Anomaly Labelling                      & Disabled         & \begin{tabular}[c]{@{}l@{}}All anomaly labelling supported \\ by ANTShapes is disabled as it \\ is not relevant to the object \\ classification task.\end{tabular}                                                                                     \\ \hline
\end{tabular}%
}

\hfill
\caption{Configuration used to generate the standard ANTShapes dataset in the simulation program.  These configurations are applied globally to all classes of objects.  ANTShapes parameters are defined as normal distributions and accept values for the mean $\mu$ and variance $\sigma$.  Per-shape parameters are sampled from these defined distributions, causing each example rendered for the dataset to be unique.}
\label{tab:antshapes_params}
\end{table}

\begin{figure}[h!]
    \centering
    \includegraphics[width=0.7\linewidth]{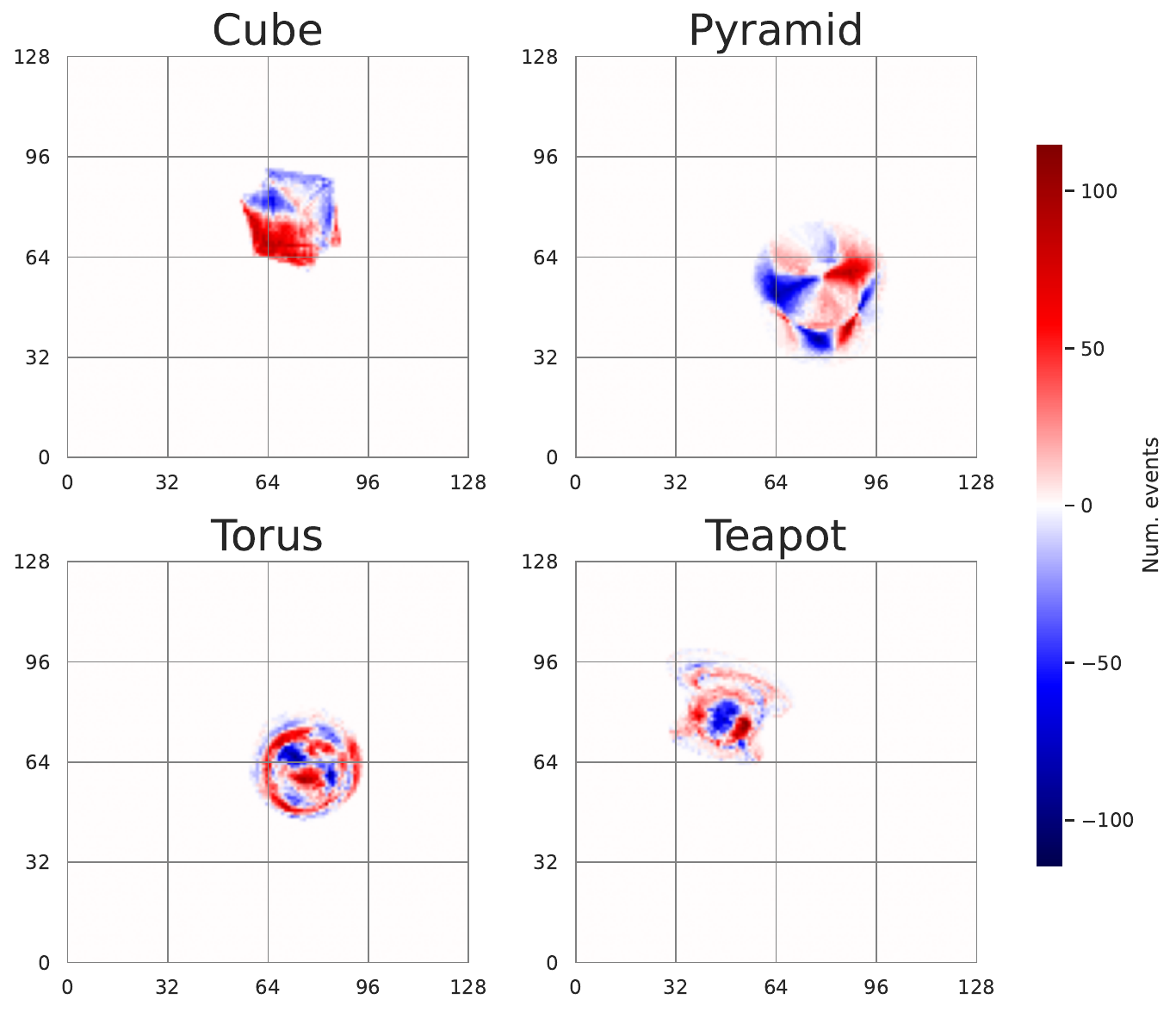}
    \caption{Examples of four classes taken from the "standard" ANTShapes dataset.  As in Figures \ref{fig:nmnist_example} to \ref{fig:dvsgesture_example}, the events pictured are accumulated over the simulation period leading to the "smearing" effect seen.  Variance in the positioning of the objects is seen between examples, as each appears in a different location in the scene.  The simulated data is also free from background noise.}
    \label{fig:antshapes_standard}
\end{figure}

The standard ANTShapes object classification dataset features a single object, selected at random from the acceptable list of classes.
The object is allowed to rotate around its centre point along all 3 axes.
A 3D unit vector with the components specified as $(x, y, z) = (1, 1, 1)$ defines a constant rotation for all simulated objects.
An example of classes contained in this dataset are shown in Figure \ref{fig:antshapes_standard}.

\subsection{Translation of Shapes}

\begin{figure}[h!]
    \centering
    \includegraphics[width=0.7\linewidth]{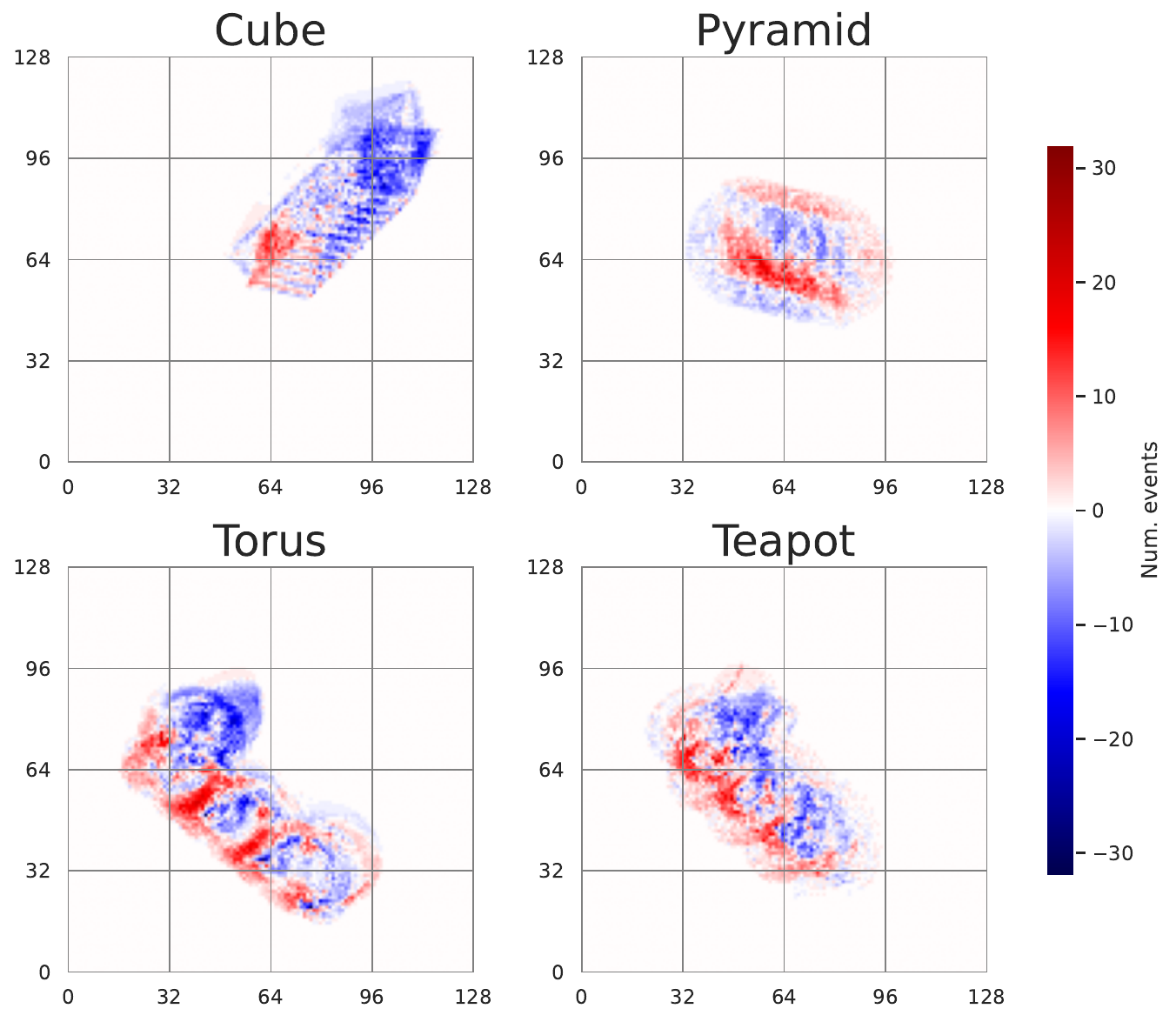}
    \caption{Examples from the ANTShapes "translation" dataset.  Compared to the standard dataset shown in Figure \ref{fig:antshapes_standard}, it can be seen that the shapes move with different trajectories and rebound from the edges of the simulated scene whilst rotating.}
    \label{fig:antshapes_translation}
\end{figure}

The methodology used to produce the standard ANTShapes dataset is repeated, with extra provisions to allow the simulated objects to move around the scene.
Four examples from the \say{translation} dataset are shown in Figure \ref{fig:antshapes_translation}.

Movement is constrained to the $X$ and $Y$ axes (left/right and up/down); objects cannot approach or recede from the camera position.
When an object reaches the boundary of the scene, it is reflected back into the space to prevent it from leaving the simulated area.
The variance of relevant global translation settings in ANTShapes were set to 0.15.
This caused most of the simulated objects in scenes to move around at a reasonable speed on average; not so fast that the events produced are aggressively smeared across time and not so slow that motion is imperceptible.
However, the normal distribution sampling method used to sample ANTShapes object parameters does allow for some extreme cases to be included in the dataset by design.
Therefore, whilst most rendered examples express steady translation vectors, some examples are allowed to move at very fast or slow speeds which increases the difficulty of the class separation task further.

\subsection{Distorted Shapes}

\begin{figure}[h!]
    \centering
    \includegraphics[width=0.7\linewidth]{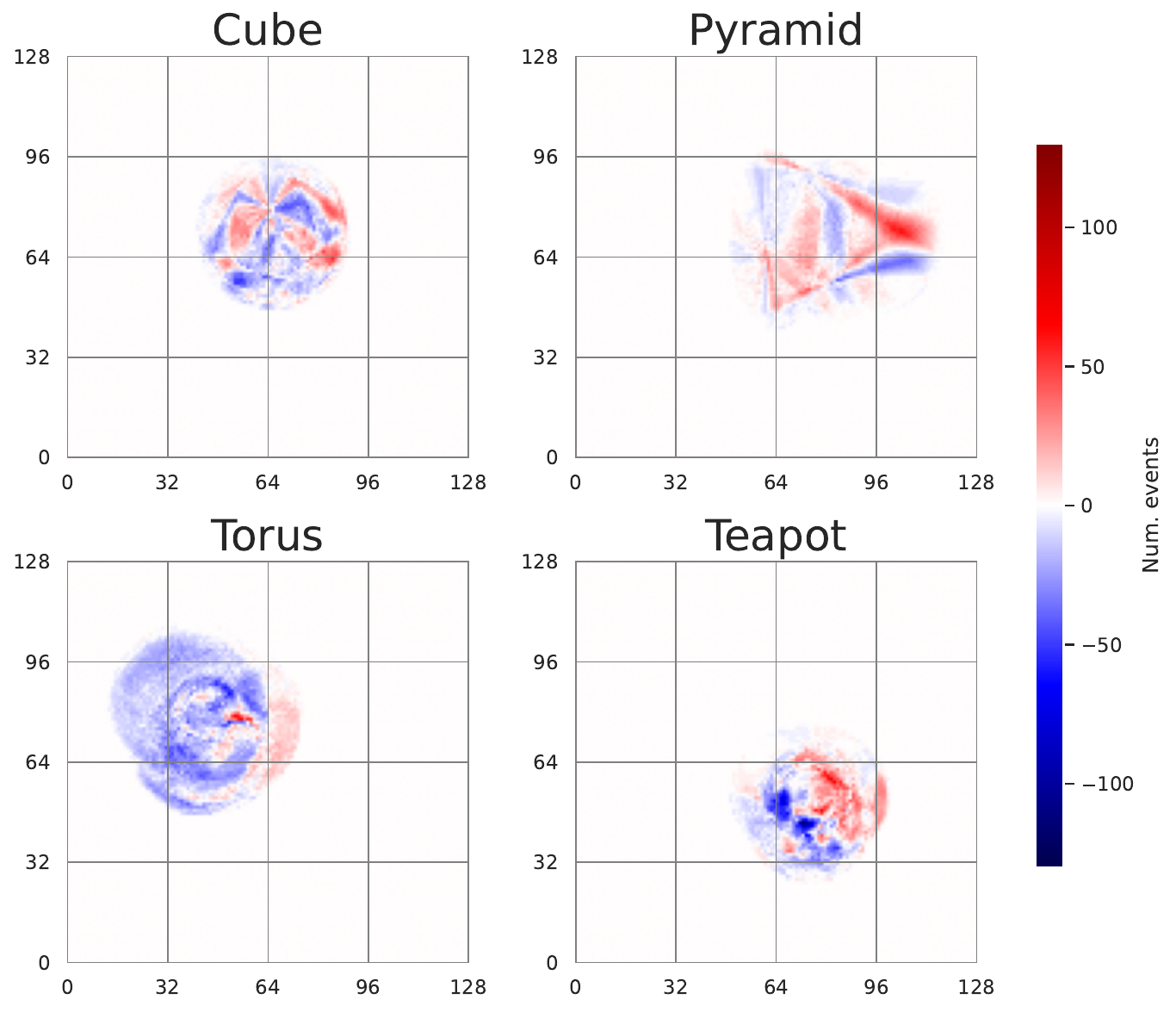}
    \caption{Four classes taken from the "distortion" dataset.  These objects each have 3 randomly-assigned scaling vectors which are each applied to the $X$, $Y$ and $Z$ dimensions to distort the object's event signature.}
    \label{fig:antshapes_distortion}
\end{figure}

A similar approach is taken to distort the scale of rendered objects.
The standard ANTShapes dataset is adapted again to introduce random scaling vectors across all 3 axes.
The variance of relevant global scaling parameters were set to 0.1.
This causes shapes to appear squashed or stretched along each axis which, when combined, results in objects appearing distorted.
Some objects appear subtly distorted whereas others are extremely squashed and stretched.
An example of this is shown in Figure \ref{fig:antshapes_distortion}.

\clearpage

\subsection{Random Rotation Vectors}

\begin{figure}[h!]
    \centering
    \includegraphics[width=0.7\linewidth]{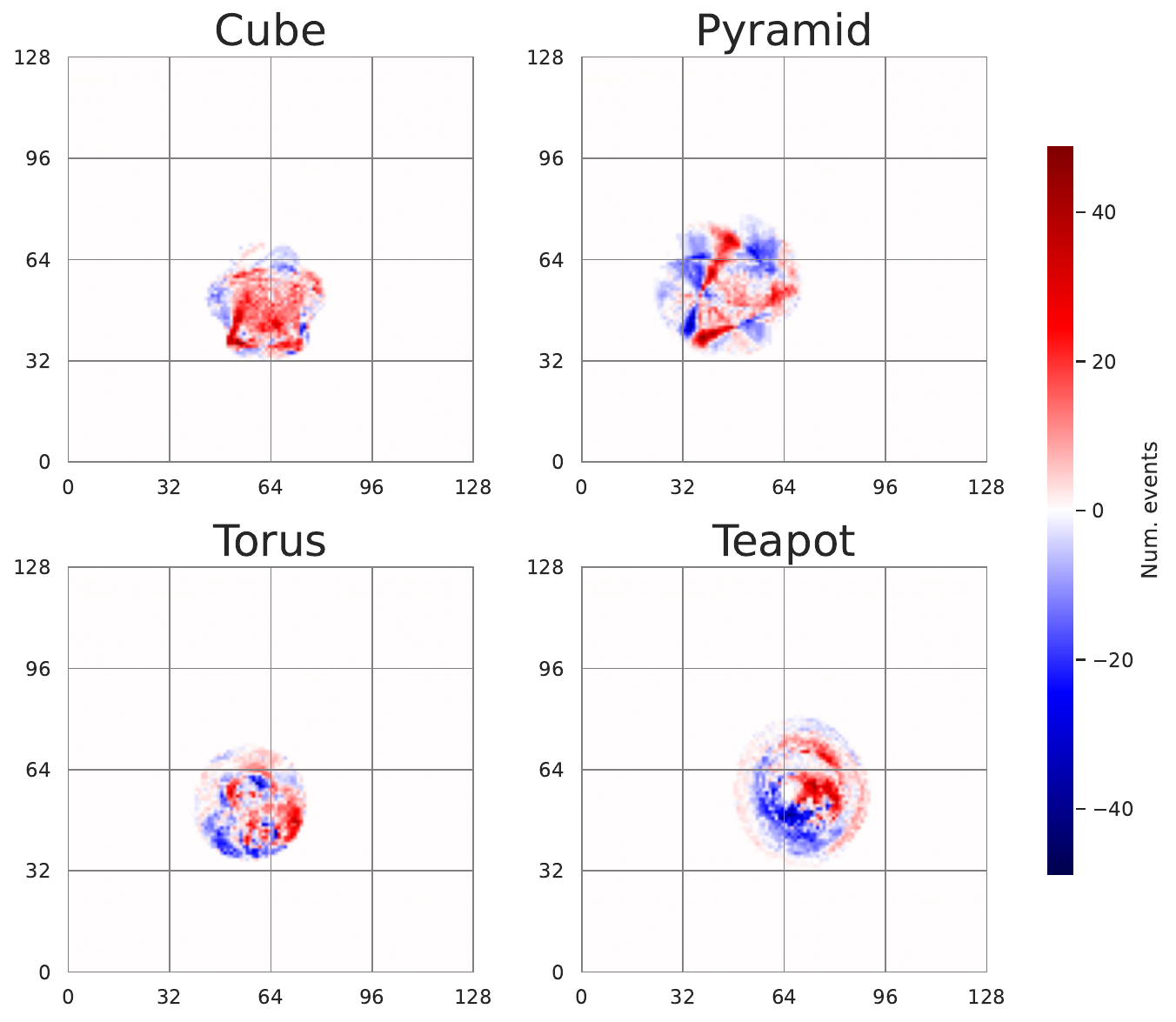}
    \caption{Four examples from the "rotation" dataset.  Unlike the standard dataset, where each object is assigned a random initial rotation then continues to rotate around the $X$, $Y$ and $Z$ axes uniformly, each simulated object is assigned a random 3D rotation vector.  This introduces additional variance between event signatures within classes by introducing independent rotational behaviour for all shapes between examples.}
    \label{fig:antshapes_rotation}
\end{figure}

In this final variation of the ANTShapes benchmarking datasets presented here, the unit rotation vector is substituted for random rotation vectors per object.
An example is shown in Figure~\ref{fig:antshapes_rotation}.

In effect, each rendered example will feature an object rotating in a random direction and at variable speeds around its central point.
It is therefore not guaranteed that an object will complete exactly 4 rotations around its centre point in the 836~ms exposure window.

To ensure that all examples expressed some meaningful rotation, two separate sets of data containing 1,500 examples each were rendered and merged.
One set used a vector of $(x, y, z) = (1, 1, 1)$ as the mean rotation speed vector, with the variance of all 3 dimensions set to 0.1 to produce variation in rotation speed and angle.
The same was repeated with a mean rotation vector of $(-1, -1, -1)$.
This causes the objects to rotate in the opposite direction, with the same variance in direction and speed.

If the dataset were to be created in a single pass with a mean rotation vector of $(0, 0, 0)$, the normal distribution sampling method in ANTShapes would cause many rendered objects to appear stationary or rotate too slowly to produce meaningful event data.

\section{Evaluation}

To validate the simulated ANTShapes datasets, they are benchmarked against the four datasets explored in Section \ref{ss:background}; N-MNIST, CIFAR10-DVS, POKER-DVS and DVSGesture.
This is achieved by constructing deep convolutional SNNs for multi-class classification and training them on the above datasets, plus the ANTShapes datasets, over multiple independent trials.
The trained networks are evaluated on out-of-sample testing data and the average performance and variance between trials is compared.
This section describes this process, using the Python \texttt{snnTorch} library for model implementation and Tonic library for dataset management.

\subsection{Dataset Preparation}
\label{ss:dataset_preparation}

Prior to constructing and training the SNNs, the datasets must be prepared and loaded into memory.
The dataset preparation procedure is as follows:

\begin{itemize}
    \item \textbf{Spatial cropping:}  The spatial resolution of all datasets are constrained to being powers of two in both axes.
    In practice, only N-MNIST and POKER-DVS are cropped (from $34 \times 34$ and $35 \times 35$ respectively, both to $32 \times 32$).  Both datasets are close to the cropped resolution making the data dropped minimal.  The rationale for this procedure is justified later in Section \ref{ss:snn_design}.
    \item \textbf{Noise suppression:}  Datasets captured using DVS cameras are cleaned of the noise that is typically introduced by these sensors.  This is implemented by filtering events which do not have an adjacent neighbouring event in space or time (that is, the event appears as spurious noise with no relation to the features of the dataset).  Noise suppression is not applied to the ANTShapes datasets as the software event simulator does not seek to emulate a DVS; the datasets are already clean of any background noise that could detract from training.
    \item \textbf{Train/test splitting:}  The 3,000 samples in each ANTShapes dataset are split into training and testing subsets.  An 80/20\% train/test split is used, producing 2,400 training samples and 600 testing samples.  These subsets are saved for later recall so the same training and testing data can be used between trials.  POKER-DVS does not define separate training and testing datasets, so the same procedure is applied to generate the training and testing sets.
    \item \textbf{PyTorch dataloader creation:}  Datasets are then loaded into \texttt{PyTorch} dataloader classes for compatibility with the \texttt{snnTorch} package.  This also allows the order of samples in training and testing sets to be shuffled prior to each training cycle.
\end{itemize}

\subsection{SNN Design}
\label{ss:snn_design}

\begin{figure}[h!]
    \centering
    \makebox[\linewidth][c]{%
        \includegraphics[width=1.2\linewidth]{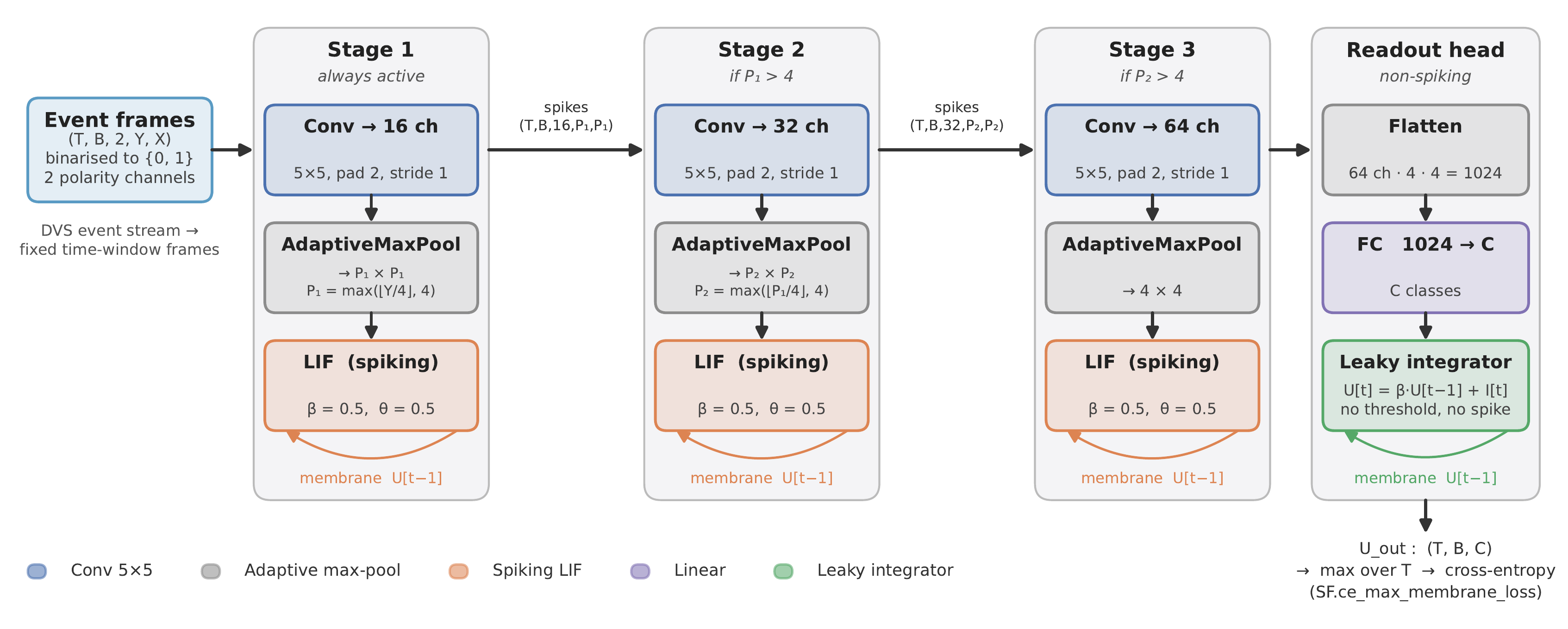}
    }
    \caption{A diagram of the spiking convolutional network design used for the following benchmarking experiments.
    Here, $T$ is the total number of time steps in the dataset where the time between frames is 1ms.  
    $X$ and $Y$ are the sensor dimensions in pixels; for ANTShapes datasets, this is the spatial resolution of the simulated scene.
    $C$ is the number of classes in the dataset being evaluated and $B$ is the batch size.
    The network architecture pools the features from previous layers, learning the most relevant features through surrogate gradient training.
    The readout head is trained on the flattened features from the output of the final hidden layer using the cross-entropy loss function.
    The stage 2 and 3 layers are optional and are only created if the length of the shortest spatial dimension $P_n$ is greater than 4 pixels.}
    \label{fig:snn_design}
\end{figure}

A convolutional SNN design is implemented for the following benchmarking experiments, a diagram of which is shown in Figure \ref{fig:snn_design}.

Working backwards from the final spiking layer towards the input, the convolutional layers reduce the feature space by a factor of 2 per layer, until the source input spatial dimensionality is reached.
For example, at $128 \times 128$ for the DVSGesture dataset, the convolutional SNN contains 3 hidden layers; first reducing the dimensionality to $64 \times 64$, then $32 \times 32$ and finally $16 \times 16$.
The number of hidden layers is therefore dependent on the spatial resolution of the input data.
Consequently, lower resolution datasets (N-MNIST and POKER-DVS specifically) contain just one hidden layer.

The convolutional SNN design is implemented such that the final layer of neurons will always be fixed into a $4 \times 4$ grid.  
Classification is performed using a densely-connected linear regression layer after the final spiking layer.
This layer is the only part of the design which is non-spiking and operates on features extracted from neuron membrane potentials which are recorded at every time step.
The features extracted are the mean, variance, minimum, maximum and range, resulting in $5 \times 16 = 80$ features per neuron for regression.
The size of the final spiking layer is therefore restricted to ensure all networks have the same feature space dimensionality for fair comparisons to be made, regardless of input data dimensionality.

A leaky integrate-and-fire (LIF) neuron design is used for all spiking neurons in the network.
Here, $\tau_m$ is the membrane time constant, $V$ is the active membrane potential, $V_{\text{rest}}$ is the resting membrane potential, $V_{\text{reset}}$ is the membrane potential after firing and $s(t)$ are the input spikes at time $t$.
Each incoming spike from presynaptic neurons increases the active voltage $V$ by nV.
If $V$ crosses the firing threshold voltage $V_{\text{th}}$, an action potential is emitted and the neuron fires.
The neuron then resets to its resting voltage $V_{\text{rest}}$.
This process is shown in Eq. \ref{eq:lif}

\begin{equation}
    \label{eq:lif}
    \begin{cases}
        \tau_m \dfrac{dV}{dt} = -\left(V - V_{\text{rest}}\right) + s(t), \\[6pt]
        V \leftarrow V_{\text{reset}} & \text{if } V \geq V_{\text{th}}.
    \end{cases}
\end{equation}

When using the \texttt{snnTorch} package for constructing SNNs, Eq. \ref{eq:lif} becomes transformed to a discrete time representation.
Here, $U$ is analogous to the neuron voltage, $U_\text{th}$ the threshold voltage, $\beta = e^{\Delta t / \tau_m}$ is a dimensionless unit representing decay rate, $X$ is the series of presynaptic spikes arriving at the neuron, $S$ is the series of spikes leaving the neuron and $W$ is the learnable weight.
This is shown in Eq. \ref{eq:lif_snntorch}.

\begin{equation}
    \label{eq:lif_snntorch}
    U[t+1] = \beta U[t] + W X[t+1] - S[t] U_{\text{th}}
\end{equation}

For the purposes of the experiments presented here, $U_{\text{th}} = 0.5$ and $\beta = 0.5$.
By extension, $\tau_m = \frac{-\Delta t}{\ln \beta} = 1.44$ms, given that each discrete time step is 1ms in length.

Classification is performed by extracting the final membrane potentials at the output layer.
Here, a fully-connected linear layer projects the features extracted from the stage 3 hidden layer to $C$-many neurons, with a neuron representing each class.
These neurons are leaky integrators; they do not ever reach a threshold and emit an action potential to postsynaptic neurons but hold a charge by integrating presynaptic spikes and allow it to leak over time.
The neuron with the highest membrane capacitance at the end of the simulation period represents the class label that is selected as the prediction in a winner-takes-all regime.
The formula for the leaky integrator neurons is shown in Eq. \ref{eq:leaky_integrator}; note the missing leaky term $-(V - V_{\text{rest}})$.

\begin{equation}
    \label{eq:leaky_integrator}
    \tau_m \frac{dV}{dt} = -V + s(t)
\end{equation}

The discrete-time version used in \texttt{snnTorch} is given in Eq. \ref{eq:leaky_integrator_snntorch}.
It is implemented in code by setting the reset mechanism for a neuron to \say{none}.

\begin{equation}
    \label{eq:leaky_integrator_snntorch}
    U[t+1] = \beta U[t] + W X[t+1]
\end{equation}

\subsection{SNN Training}

The convolutional SNN is trained using the \texttt{snnTorch} surrogate gradient process~\cite{eshraghian2023snntorch}. 
The derivative of the spike emission function is the Dirac delta, $\partial S / \partial U = \delta(U - U_{\text{th}})$, which is zero everywhere except at threshold, where it is undefined. 
As this factor enters multiplicatively into every credit-assignment path, the exact gradient is unavailable for optimisation.

Backpropagation through time unrolls Eq.~\ref{eq:lif_snntorch} over the $T$ simulation steps, so that the loss gradient accumulates along both a spatial path through the weights and a temporal path through the membrane state \cite{eshraghian2023snntorch}, as in Eq.~\ref{eq:bptt}.
The reset term $\mathrm{d}U[t]/\mathrm{d}W$ is shown to the right of the equation.

\begin{equation}
    \frac{\partial \mathcal{L}}{\partial W}
    = \sum_{t=1}^{T}
      \frac{\partial \mathcal{L}}{\partial S[t]}\,
      \frac{\partial S[t]}{\partial U[t]}\,
      \frac{\mathrm{d} U[t]}{\mathrm{d} W},
    \qquad
    \frac{\mathrm{d} U[t]}{\mathrm{d} W} = \sum_{s=1}^{t} \beta^{\,t-s} X[s]
    \label{eq:bptt}
\end{equation}

As \texttt{snnTorch} detaches the reset term of Eq.~\ref{eq:lif_snntorch} from the computational graph, the recurrent Jacobian collapses to the scalar decay rate, $\partial U[t] / \partial U[t-1] = \beta$ \cite{eshraghian2023snntorch}.
Unrolling the resulting recursion $\mathrm{d}U[t]/\mathrm{d}W = X[t] + \beta\,\mathrm{d}U[t-1]/\mathrm{d}W$ weights the input at time step $t-k$ by $\beta^{k}$ so credit assignment decays geometrically over time.

The surrogate gradient method leaves the forward pass unchanged~\cite{neftci2019surrogate} and binary spikes are still emitted according to Eq.~\ref{eq:lif_snntorch}. 
Only in the backward pass is the Dirac delta function replaced by the derivative of a smooth relaxation $\sigma(x)$ of the emission function, allowing gradients to propagate.
This is shown in Eq.~\ref{eq:surrogate}.

\begin{equation}
    \frac{\partial S[t]}{\partial U[t]} := \sigma'\!\left(U[t] - U_{\text{th}}\right)
    \label{eq:surrogate}
\end{equation}

The default \texttt{snnTorch} surrogate, used throughout the forthcoming experiments, is the shifted arctangent with $\alpha = 2$~\cite{fang2021plif}, given in Eq.~\ref{eq:atan}.

\begin{equation}
    \sigma(x) = \frac{1}{2} + \frac{1}{\pi}\arctan\!\left(\frac{\pi \alpha x}{2}\right),
    \qquad
    \sigma'(x) = \frac{\alpha/2}{1 + \left(\frac{\pi \alpha x}{2}\right)^{2}}
    \label{eq:atan}
\end{equation}

Because the forward pass emits binary spikes while the backward pass differentiates a distinct smooth function, the resulting update is a biased descent direction that approximates the true gradient.

\subsection{Benchmarking Results}

To visualise the results obtained from the SNN for the ANTShapes datasets with respect to the four established datasets, plots have been produced and included at the end of this paper for legibility (Section \ref{ss:apx}).
Table \ref{tab:results_overview} presents an overview of the mean accuracy and standard deviation for all datasets evaluated.

\begin{table}[h!]
\resizebox{\columnwidth}{!}{%
\begin{tabular}{|l|l|l|l|l|l|l|l|l|}
\hline
              & \textbf{N-MNIST} & \textbf{CIFAR10-DVS} & \textbf{POKER-DVS} & \textbf{DVSGesture} & \textbf{\begin{tabular}[c]{@{}l@{}}ANTShapes\\ (Standard)\end{tabular}} & \textbf{\begin{tabular}[c]{@{}l@{}}ANTShapes\\ (Translation)\end{tabular}} & \textbf{\begin{tabular}[c]{@{}l@{}}ANTShapes\\ (Distortion)\end{tabular}} & \textbf{\begin{tabular}[c]{@{}l@{}}ANTShapes\\ (Rotation)\end{tabular}} \\ \hline
Mean Accuracy & 97\%             & 55.6\%               & 96.6\%             & 88.9\%              & 92.2\%                                                                  & 89.1\%                                                                     & 83.5\%                                                                    & 89.8\%                                                                  \\ \hline
Variance &
$2.8 \times 10^{-3}$ &
$5.3 \times 10^{-3}$ &
$11.0 \times 10^{-3}$ &
$12.0 \times 10^{-3}$ &
$12.0 \times 10^{-3}$ &
$8.6 \times 10^{-3}$ &
$16.0 \times 10^{-3}$ &
$7.8 \times 10^{-3}$ \\ \hline
\end{tabular}%
}
\hfill
\caption{Summary of the mean accuracy and variance for all datasets.  10 convolutional SNNs, each with unique random initial weights, were trained.  The mean accuracy and variance of the ensembles are shown.}
\label{tab:results_overview}
\end{table}

As expected, the standard ANTShapes dataset achieved the highest prediction accuracy among all four of the ANTShapes datasets evaluated at 92.2\% mean accuracy.
This is because it represents the simplest classification problem of the ANTShapes datasets, insofar as the examples represented within it do not exhibit any variance in rotational or translational behaviour and are not distorted in shape.
The distorted shape dataset achieved the lowest reported mean accuracy at 83.5\%.
This is also to be expected given the deviations in visual signatures between examples of the same class in this dataset.
The other ANTShapes datasets used objects that are uniform in shape across all examples which allowed for simpler classification, despite deviations in rotation angle and speed or translation trajectory.

\begin{figure}[h!]
    \centering
    \makebox[\linewidth][c]{%
        \includegraphics[width=1.2\linewidth]{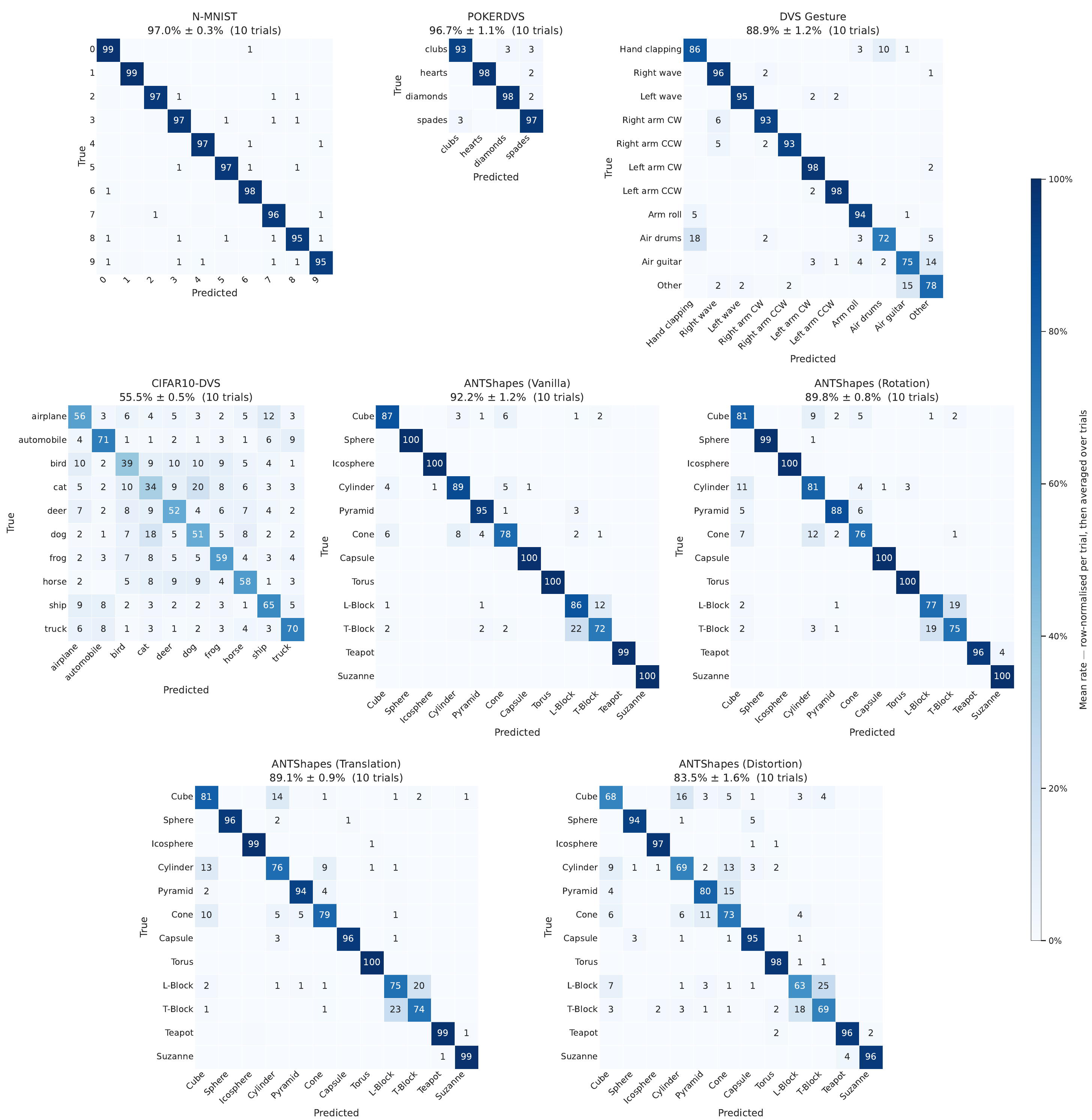}
    }
    \caption{Confusion matrices produced for each dataset, illustrating the class confusions predicted by the convolutional SNN.
    The mean over all 10 trials is plotted here to give an impression of the average confusion matrix produced by the network for a given dataset.}
    \label{fig:confusion_grid}
\end{figure}

The classification performance of the four ANTShapes datasets are largely consistent with the established N-MNIST, POKER-DVS and DVSGesture datasets, each achieving comparable mean accuracy with respect to the novel datasets.
This observation is reinforced for the confusion matrices produced for the ANT-Shapes datasets, shown in Figure \ref{fig:confusion_grid}.
The clear diagonal shown in these plots demonstrates that class separation has been consistently achieved by the SNN for all examples.
However, some meaningful confusion exists between the L-block and T-block classes in all examples.
This is due to the visual similarity between these two classes, which is confirmed by the confusion matrices produced for the ANTShapes rotation and distortion datasets.
These datasets introduce additional confusions between these classes by representing only partial rotations or by distorting the proportions of the objects so their event vision signatures appear more similar to each other under certain conditions.

A clear outlier is the CIFAR10-DVS dataset, which achieved a mean accuracy of 55.6\%.
One possible reason for this is the domination of positive-polarity spiking events, as visualised in Figure \ref{fig:cifar10dvs_example}.
This uneven distribution of events is not present in any of the other datasets and is the most likely source of the class confusions for this reason.
Examining the Principal Component Analysis (PCA) cluster diagram for the CIFAR10-DVS dataset in Figure \ref{fig:pca_grid} reveals significant overlap between the 10 classes, suggesting strong similarities between their features which are difficult for the SNN to separate.
Further comparative evidence can be seen in the confusion matrix diagrams presented in Figure \ref{fig:confusion_grid}, where the number of class confusions for the CIFAR10-DVS dataset are visibly higher than any other evaluated dataset.
Despite this, there is still a clear diagonal visible in the confusion matrix, demonstrating the ability of the convolutional SNN to distinguish between classes in some cases.

\subsection{Discussion}

\begin{figure}[h!]
    \centering
    \makebox[\linewidth][c]{%
        \includegraphics[trim={0 0 0 2cm},clip,width=1.2\linewidth]{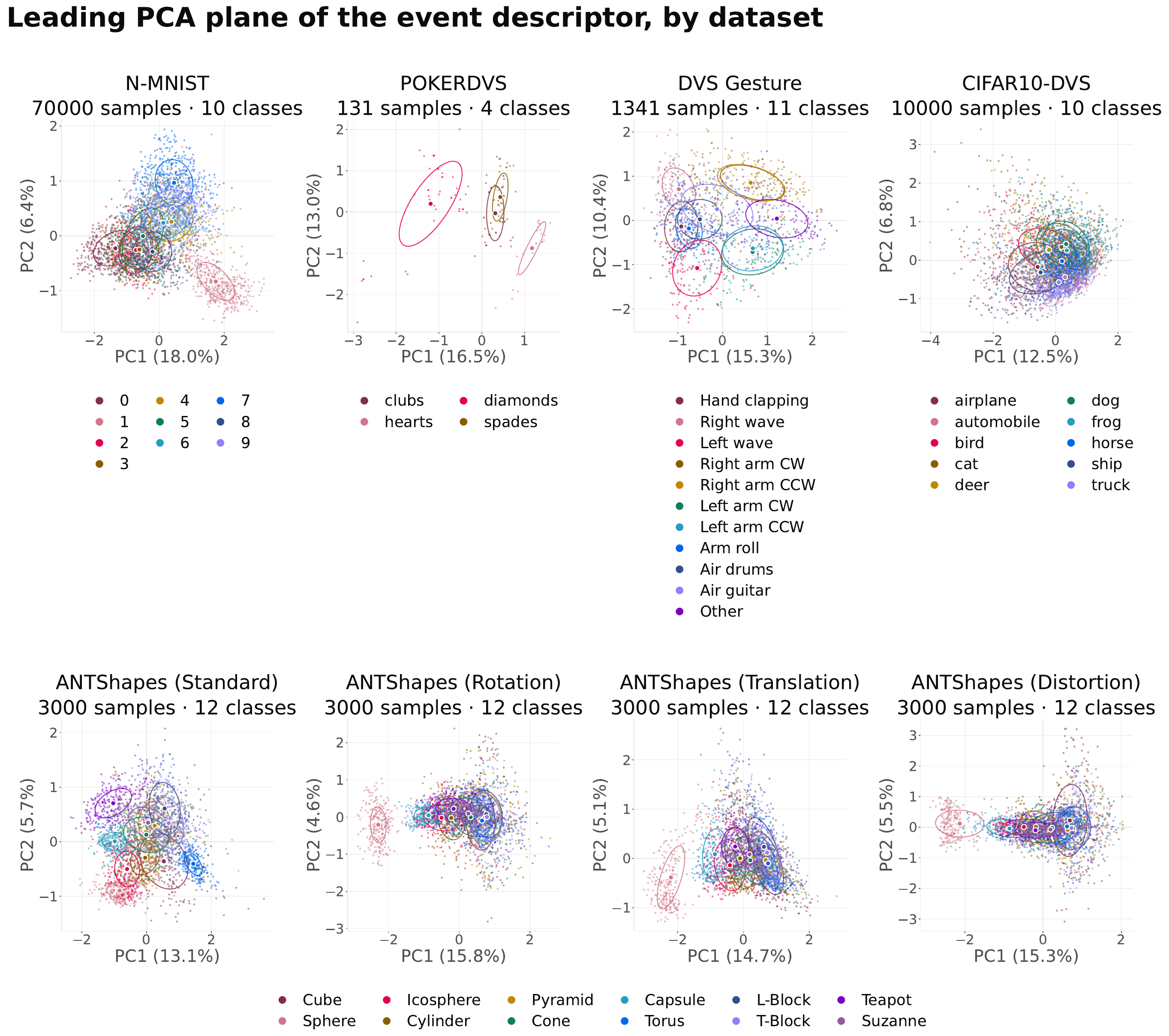}
    }
    \caption{PCA plots created for each of the datasets evaluated in this paper.
    The DVSGesture plots show clear separation, but also class overlap between clockwise (CW) and counter-clockwise (CCW) motion, which leads to difficulties in separating the classes.
    POKER-DVS shows clear separation, but only contains 4 classes and a limited number of samples.
    CIFAR10-DVS shows extreme overlap of principal elements for all classes, demonstrating that the principal components for each class are largely similar and are difficult to separate.
    The standard ANTShapes dataset has clear separation between most classes, with some overlap between the L-block and T-block objects, and between the various smooth-faced objects (such as cones and capsules).
    The separation between principal component clusters becomes less well-defined when random rotation, translation and scaling (distortion) vectors are introduced to the objects presented in each example.
    For all ANTShapes datasets, the sphere objects remain the most cleanly separated from the rest of the classes.
    This is due to their near-uniform appearance when rotated, with events only being generated from the motion of simulated roughness on the sphere's surface.
    Their event signature is therefore distinct from the other objects presented, which can appear different depending on their angle of rotation.}
    \label{fig:pca_grid}
\end{figure}

The outcome of the experiments presented in this paper are summarised as follows:

\begin{itemize}
    \item \textbf{ANTShapes datasets are fit for purpose:}  The standard ANTShapes dataset provides a clear baseline for event-based object recognition tasks.  Besides the class of object, variance between examples exists as independent initial rotation vectors and positions in the scene.  The other three datasets produced (\say{translation}, \say{distortion} and \say{rotation}) each provide an additional challenge by introducing additional variance between examples.  This is confirmed by the results shown in Table \ref{tab:results_overview}, where the standard dataset scored higher than the other ANTShapes datasets.
    \item \textbf{ANTShapes software can generate DVS-like datasets:}  The four ANT-Shapes datasets shown in this paper each have large number of classes and examples and are rendered at a high resolution, all relative to the N-MNIST, CIFAR10-DVS, POKER-DVS and DVSGesture datasets also investigated here.  Through the various visualisations and experimental procedures shown in the preceeding sections, it can be confirmed that the ANTShapes simulation tool can produce data suitable for event-based computer vision tasks.
    \item \textbf{Separation of class features in established datasets may be insufficient for object classification experiments:}  It was also revealed through plotting the principal components of individual dataset examples in cluster diagrams that the CIFAR10-DVS dataset exhibits highly similar features between classes (Figure \ref{fig:pca_grid}).
    As demonstrated in the experiments conducted here, this overlap of features leads to difficulty in classification.  This may complicate the design and analysis of SNNs by researchers if the data is not suitable for the task at hand.
\end{itemize}

Some limitations in the experimental process can also be identified however.
Firstly, the datasets are evaluated only using the spiking convolutional network design described in Section \ref{ss:snn_design}.
Although care has been taken to assess the average SNN performance over a number of trials, other network models could be used to assess the viability of the datasets for event-based computer vision tasks.

Furthermore, there is not a significant degradation in performance between the standard ANTShapes datasets and the three more challenging datasets; between the standard and distortion datasets, there is only a $<10\%$ loss in average classification performance.
Therefore, the additional challenge presented by these datasets may not be sufficient for future experiments and more sophisticated examples may need to be simulated.

Finally, the PCA plots shown in Figure \ref{fig:pca_grid} do not fully describe the separation of class features.
The clustering does indicate the location of principal features for all examples in each dataset discussed, which gives a strong impression of class separation.
However, the event data being separated by the SNN is extremely high-dimensional and can only be represented in two dimensions with an accepted loss in accuracy with respect to the true features.

\section{Conclusions}

To conclude, the four novel ANTShapes datasets produced for this research fulfil their intended purposes for object classification.
This has been validated by benchmarking them against existing event-based datasets for object classification, where each achieved comparable results to existing datasets that are used as approximations for object classification.
The datasets presented here encode spatiotemporal changes in simple, idealised scenes containing single objects which are free from the noise captured by real-world event-based sensors.
This makes them ideal candidates for event-based computer vision experiments which rely on the classification of 3D objects.
Furthermore, the experiments demonstrated here validate the data simulated by the ANTShapes simulation tool and demonstrates its suitability for simulating data specialised towards event-based computer vision tasks.

\subsection{Further Work}

As stated in the introduction to this paper, object classification is the fundamental core problem from which deeper event-based scene decomposition, object recognition and anomaly detection pipelines can be built.
To this end, the ANTShapes simulation tool is capable of automatically simulating and labelling scenes containing multiple objects.
Furthermore, the tool automatically labels \say{anomalous} objects as defined by their deviations from expected \say{normal} behaviours based on central limit theorem.
This describes a fuzzy boundary between normal and anomalous objects as defined by their behaviours, which can be exaggerated by the inclusion of multiple behaviour parameters (such as translation and rotation vectors).

Therefore, ANTShapes is capable of simulating rich 3D scenes with automated labelling of anomalies through pixel-perfect masks and bounding boxes.
Further work in the wider context of autonomous anomaly detection could use the tool to simulate more detailed scenes, consisting of multiple objects.
One suggested research opportunity is scene decomposition; decomposing an event-based scene containing many objects into \say{regions of interest} encapsulating single objects.
Algorithms have been proposed to decompose an event-based scene, which could be applied to data simulated by ANTShapes.
From here, the object classification approach presented in this paper could be used to classify the objects present in the scene.

A further limitation of the datasets presented here is that the classification task is simple overall.
The classification of shapes is time invariant; the class of shape cannot change in a single example, so only one class label must be predicted that describes the shape shown in the scene.
A more complex, time dependent problem could involve classifying more than one type of shape per example and sensing when the object is replaced by one of another class.
This would be beneficial for assessing the ability of an SNN to separate temporal events by incorporating these features into the classification task.
In such a scenario, the SNN would be tasked with classifying the shapes in the scene and the onset of change, where one object is switched for another.
Therefore, whilst the four ANTShapes datasets presented here contain spatiotemporal events that describe the changing content of a scene, the classes described in each example do not change over time.

\section{Data Availability}

The ANTShapes simulation tool is available on GitHub: \url{https://github.com/EDGYOrganism/ANTShapes/}.

\hfill

The benchmarking datasets produced for these experiments are available on Zenodo~\cite{middleton2026dataset}: \url{https://zenodo.org/records/21813140}.

\begin{credits}

\subsubsection{\ackname} 

This work was funded by EPSRC grant number EP/Y030133/1. The work was also supported through the NimbleAI project, funded via the Horizon Europe Research and Innovation programme (Grant Agreement 101070679), and UKRI under the UK government’s Horizon Europe funding guarantee (Grant Agreement 10039070).

\subsubsection{\discintname}

There are no competing interests from the authors.

\end{credits}

\clearpage

\section{Additional Figures}
\label{ss:apx}

\clearpage

\clearpage

\bibliographystyle{splncs04unsrt}
\bibliography{refs}

@article{davies2018loihi,
  author  = {Davies, Mike and Srinivasa, Narayan and Lin, Tsung-Han and
             Chinya, Gautham and Cao, Yongqiang and Choday, Sri Harsha and
             Dimou, Georgios and Joshi, Prasad and Imam, Nabil and Jain, Shweta and
             Liao, Yuyun and Lin, Chit-Kwan and Lines, Andrew and Liu, Ruokun and
             Mathaikutty, Deepak and McCoy, Steven and Paul, Arnab and Tse, Jonathan and
             Venkataramanan, Guruguhanathan and Weng, Yi-Hsin and Wild, Andreas and
             Yang, Yoonseok and Wang, Hong},
  title   = {Loihi: A Neuromorphic Manycore Processor with On-Chip Learning},
  journal = {IEEE Micro},
  volume  = {38},
  number  = {1},
  pages   = {82--99},
  year    = {2018},
  doi     = {10.1109/MM.2018.112130359}
}

@article{painkras2013spinnaker,
  author  = {Painkras, Eustace and Plana, Luis A. and Garside, Jim and
             Temple, Steve and Galluppi, Francesco and Patterson, Cameron and
             Lester, David R. and Brown, Andrew D. and Furber, Steve B.},
  title   = {{SpiNNaker}: A 1-{W} 18-Core System-on-Chip for
             Massively-Parallel Neural Network Simulation},
  journal = {IEEE Journal of Solid-State Circuits},
  volume  = {48},
  number  = {8},
  pages   = {1943--1953},
  year    = {2013},
  doi     = {10.1109/JSSC.2013.2259038}
}

@misc{schemmel2020brainscales,
  author        = {Schemmel, Johannes and Billaudelle, Sebastian and
                   Dauer, Phillip and Weis, Johannes},
  title         = {Accelerated Analog Neuromorphic Computing},
  year          = {2020},
  eprint        = {2003.11996},
  archivePrefix = {arXiv},
  primaryClass  = {cs.NE},
  note          = {arXiv:2003.11996},
  doi           = {10.48550/arXiv.2003.11996}
}

@article{indiveri2015memoryandinformationprocessing,
  author  = {Indiveri, Giacomo and Liu, Shih-Chii},
  title   = {Memory and Information Processing in Neuromorphic Systems},
  journal = {Proceedings of the IEEE},
  volume  = {103},
  number  = {8},
  pages   = {1379--1397},
  year    = {2015},
  doi     = {10.1109/JPROC.2015.2444094}
}

@article{kudithipudi2025neuromorphiccomputingatscale,
  author  = {Kudithipudi, Dhireesha and Schuman, Catherine and
             Vineyard, Craig M. and Pandit, Tej and Merkel, Cory and
             Kubendran, Rajkumar and Aimone, James B. and Orchard, Garrick and
             Mayr, Christian and Benosman, Ryad and Hays, Joe and Young, Cliff and
             Bartolozzi, Chiara and Majumdar, Amitava and Cardwell, Suma George and
             Payvand, Melika and Buckley, Sonia and Kulkarni, Shruti and
             Gonzalez, Hector A. and Cauwenberghs, Gert and Thakur, Chetan Singh and
             Subramoney, Anand and Furber, Steve},
  title   = {Neuromorphic Computing at Scale},
  journal = {Nature},
  volume  = {637},
  number  = {8047},
  pages   = {801--812},
  year    = {2025},
  doi     = {10.1038/s41586-024-08253-8}
}

@misc{cheng2026braininspiredaiedgeintelligence,
  author        = {Cheng, Yingchao and Wang, Meijia and Hao, Zhifeng and
                   Buyya, Rajkumar},
  title         = {Brain-Inspired {AI} for Edge Intelligence: A Systematic Review},
  year          = {2026},
  eprint        = {2603.26722},
  archivePrefix = {arXiv},
  primaryClass  = {cs.NE},
  note          = {arXiv:2603.26722},
  doi           = {10.48550/arXiv.2603.26722}
}

@inproceedings{taunyazov2020eventdrivenvisualtactilesensinglearning,
  author    = {Taunyazov, Tasbolat and Sng, Weicong and See, Hian Hian and
               Lim, Brian and Kuan, Jethro and Ansari, Abdul Fatir and
               Tee, Benjamin C. K. and Soh, Harold},
  title     = {Event-Driven Visual-Tactile Sensing and Learning for Robots},
  booktitle = {Proceedings of Robotics: Science and Systems (RSS)},
  address   = {Corvallis, Oregon, USA},
  month     = jul,
  year      = {2020},
  doi       = {10.15607/RSS.2020.XVI.020}
}

@article{cramer2022surrogategradients,
  author  = {Cramer, Benjamin and Billaudelle, Sebastian and Kanya, Simeon and
             Leibfried, Aron and Gr{\"u}bl, Andreas and Karasenko, Vitali and
             Pehle, Christian and Schreiber, Korbinian and Stradmann, Yannik and
             Weis, Johannes and Schemmel, Johannes and Zenke, Friedemann},
  title   = {Surrogate Gradients for Analog Neuromorphic Computing},
  journal = {Proceedings of the National Academy of Sciences},
  volume  = {119},
  number  = {4},
  pages   = {e2109194119},
  year    = {2022},
  doi     = {10.1073/pnas.2109194119}
}

@article{haessig2019spiking,
  author  = {Haessig, Germain and Berthelon, Xavier and Ieng, Sio-Hoi and
             Benosman, Ryad},
  title   = {A Spiking Neural Network Model of Depth from Defocus for
             Event-Based Neuromorphic Vision},
  journal = {Scientific Reports},
  volume  = {9},
  number  = {1},
  pages   = {3744},
  year    = {2019},
  doi     = {10.1038/s41598-019-40064-0}
}

@article{orchard2015converting,
  author  = {Orchard, Garrick and Jayawant, Ajinkya and Cohen, Gregory K. and
             Thakor, Nitish},
  title   = {Converting Static Image Datasets to Spiking Neuromorphic Datasets
             Using Saccades},
  journal = {Frontiers in Neuroscience},
  volume  = {9},
  pages   = {437},
  year    = {2015},
  doi     = {10.3389/fnins.2015.00437}
}

@article{li2017cifar10dvs,
  author  = {Li, Hongmin and Liu, Hanchao and Ji, Xiangyang and Li, Guoqi and
             Shi, Luping},
  title   = {{CIFAR10-DVS}: An Event-Stream Dataset for Object Classification},
  journal = {Frontiers in Neuroscience},
  volume  = {11},
  pages   = {309},
  year    = {2017},
  doi     = {10.3389/fnins.2017.00309}
}

@article{serrano2015poker,
  author  = {Serrano-Gotarredona, Teresa and Linares-Barranco, Bernab{\'e}},
  title   = {{Poker-DVS} and {MNIST-DVS}. Their History, How They Were Made,
             and Other Details},
  journal = {Frontiers in Neuroscience},
  volume  = {9},
  pages   = {481},
  year    = {2015},
  doi     = {10.3389/fnins.2015.00481}
}

@inproceedings{amir2017low,
  author    = {Amir, Arnon and Taba, Brian and Berg, David and Melano, Timothy and
               McKinstry, Jeffrey and Di Nolfo, Carmelo and Nayak, Tapan and
               Andreopoulos, Alexander and Garreau, Guillaume and Mendoza, Marcela and
               Kusnitz, Jeff and Debole, Michael and Esser, Steve and Delbruck, Tobi and
               Flickner, Myron and Modha, Dharmendra},
  title     = {A Low Power, Fully Event-Based Gesture Recognition System},
  booktitle = {Proceedings of the IEEE Conference on Computer Vision and
               Pattern Recognition (CVPR)},
  pages     = {7243--7252},
  year      = {2017},
  doi       = {10.1109/CVPR.2017.781}
}

@article{brandli2014dvs,
  author  = {Brandli, Christian and Berner, Raphael and Yang, Minhao and
             Liu, Shih-Chii and Delbruck, Tobi},
  title   = {A 240 $\times$ 180 130 {dB} 3 $\mu$s Latency Global Shutter
             Spatiotemporal Vision Sensor},
  journal = {IEEE Journal of Solid-State Circuits},
  volume  = {49},
  number  = {10},
  pages   = {2333--2341},
  year    = {2014},
  doi     = {10.1109/JSSC.2014.2342715}
}

@article{eshraghian2023snntorch,
  author  = {Eshraghian, Jason K. and Ward, Max and Neftci, Emre O. and
             Wang, Xinxin and Lenz, Gregor and Dwivedi, Girish and
             Bennamoun, Mohammed and Jeong, Doo Seok and Lu, Wei D.},
  title   = {Training Spiking Neural Networks Using Lessons from Deep Learning},
  journal = {Proceedings of the IEEE},
  volume  = {111},
  number  = {9},
  pages   = {1016--1054},
  year    = {2023},
  doi     = {10.1109/JPROC.2023.3308088}
}

@article{neftci2019surrogate,
  author  = {Neftci, Emre O. and Mostafa, Hesham and Zenke, Friedemann},
  title   = {Surrogate Gradient Learning in Spiking Neural Networks: Bringing
             the Power of Gradient-Based Optimization to Spiking Neural Networks},
  journal = {IEEE Signal Processing Magazine},
  volume  = {36},
  number  = {6},
  pages   = {51--63},
  year    = {2019},
  doi     = {10.1109/MSP.2019.2931595}
}

@inproceedings{fang2021plif,
  author    = {Fang, Wei and Yu, Zhaofei and Chen, Yanqi and
               Masquelier, Timoth{\'e}e and Huang, Tiejun and Tian, Yonghong},
  title     = {Incorporating Learnable Membrane Time Constant to Enhance
               Learning of Spiking Neural Networks},
  booktitle = {Proceedings of the IEEE/CVF International Conference on
               Computer Vision (ICCV)},
  pages     = {2661--2671},
  month     = oct,
  year      = {2021},
  doi       = {10.1109/ICCV48922.2021.00266}
}

@article{maass1997networks,
  author  = {Maass, Wolfgang},
  title   = {Networks of Spiking Neurons: The Third Generation of Neural
             Network Models},
  journal = {Neural Networks},
  volume  = {10},
  number  = {9},
  pages   = {1659--1671},
  year    = {1997},
  doi     = {10.1016/S0893-6080(97)00011-7}
}

@article{roy2019spikebased,
  author  = {Roy, Kaushik and Jaiswal, Akhilesh and Panda, Priyadarshini},
  title   = {Towards Spike-Based Machine Intelligence with Neuromorphic
             Computing},
  journal = {Nature},
  volume  = {575},
  number  = {7784},
  pages   = {607--617},
  year    = {2019},
  doi     = {10.1038/s41586-019-1677-2}
}

@article{gallego2022eventbasedvision,
  author  = {Gallego, Guillermo and Delbr{\"u}ck, Tobi and Orchard, Garrick and
             Bartolozzi, Chiara and Taba, Brian and Censi, Andrea and
             Leutenegger, Stefan and Davison, Andrew J. and Conradt, J{\"o}rg and
             Daniilidis, Kostas and Scaramuzza, Davide},
  title   = {Event-Based Vision: A Survey},
  journal = {IEEE Transactions on Pattern Analysis and Machine Intelligence},
  volume  = {44},
  number  = {1},
  pages   = {154--180},
  year    = {2022},
  doi     = {10.1109/TPAMI.2020.3008413}
}

@article{iyer2021nmnist,
  author  = {Iyer, Laxmi R. and Chua, Yansong and Li, Haizhou},
  title   = {Is Neuromorphic {MNIST} Neuromorphic? {A}nalyzing the
             Discriminative Power of Neuromorphic Datasets in the Time Domain},
  journal = {Frontiers in Neuroscience},
  volume  = {15},
  pages   = {608567},
  year    = {2021},
  doi     = {10.3389/fnins.2021.608567}
}

@misc{middleton2026antshapes,
  author        = {Middleton, Mike and Ali, Teymoor and Kayan, Hakan and
                   Sen Bhattacharya, Basabdatta and Perera, Charith and
                   Rhodes, Oliver and Gheorghiu, Elena and Vousden, Mark and
                   Trefzer, Martin A.},
  title         = {Modelling and Simulation of Neuromorphic Datasets for
                   Anomaly Detection in Computer Vision},
  year          = {2026},
  eprint        = {2602.23514},
  archivePrefix = {arXiv},
  primaryClass  = {cs.CV},
  note          = {arXiv:2602.23514},
  doi           = {10.48550/arXiv.2602.23514}
}

@Article{middleton2026eventbasedvisionattheedge,
AUTHOR = {Middleton, Michael and Ali, Teymoor and Baikas, Epifanios and Kayan, Hakan and Sen Bhattacharya, Basabdatta and Gheorghiu, Elena and Vousden, Mark and Perera, Charith and Rhodes, Oliver and Trefzer, Martin A.},
TITLE = {Event-Based Vision at the Edge: A Review},
JOURNAL = {Brain Sciences},
VOLUME = {16},
YEAR = {2026},
NUMBER = {4},
ARTICLE-NUMBER = {422},
URL = {https://www.mdpi.com/2076-3425/16/4/422},
PubMedID = {42041830},
ISSN = {2076-3425},
DOI = {10.3390/brainsci16040422}
}

@misc{zhang2024accurateefficienteventbasedsemantic,
      title={Accurate and Efficient Event-based Semantic Segmentation Using Adaptive Spiking Encoder-Decoder Network}, 
      author={Rui Zhang and Luziwei Leng and Kaiwei Che and Hu Zhang and Jie Cheng and Qinghai Guo and Jiangxing Liao and Ran Cheng},
      year={2024},
      eprint={2304.11857},
      archivePrefix={arXiv},
      primaryClass={cs.CV},
      url={https://arxiv.org/abs/2304.11857}, 
}

@article{zhang2023energyefficientspikingsegmenter,
  author = {Zhang, H and Fan, X and Zhang, Y},
  year = {2023},
  month = {August},
  title = {Energy-Efficient Spiking Segmenter for Frame and Event-Based Images},
  journal = {Biomimetics},
  volume = {8},
  number = {4},
  pages = {356},
  doi = {10.3390/biomimetics8040356},
  pmid = {37622961},
  pmcid = {PMC10452323}
}

@INPROCEEDINGS{viale2022lanesnns,
  author={Viale, Alberto and Marchisio, Alberto and Martina, Maurizio and Masera, Guido and Shafique, Muhammad},
  booktitle={2022 IEEE/RSJ International Conference on Intelligent Robots and Systems (IROS)}, 
  title={LaneSNNs: Spiking Neural Networks for Lane Detection on the Loihi Neuromorphic Processor}, 
  year={2022},
  volume={},
  number={},
  pages={79-86},
  doi={10.1109/IROS47612.2022.9981034}}

@article{milde2017obstacleavoidance,
  author = {Milde, MB and Blum, H and Dietmüller, A and Sumislawska, D and Conradt, J and Indiveri, G and Sandamirskaya, Y},
  year = {2017},
  month = {July},
  title = {Obstacle avoidance and target acquisition for robot navigation using a mixed signal analog/digital neuromorphic processing system},
  journal = {Front Neurorobot},
  volume = {11},
  pages = {28},
  doi = {10.3389/fnbot.2017.00028},
  pmid = {28747883},
  pmcid = {PMC5507184}
}

@inproceedings{qian2025ucfcrimedvs,
author = {Qian, Yuanbin and Ye, Shuhan and Wang, Chong and Cai, Xiaojie and Qian, Jiangbo and Wu, Jiafei},
title = {UCF-Crime-DVS: a novel event-based dataset for video anomaly detection with spiking neural networks},
year = {2025},
isbn = {978-1-57735-897-8},
publisher = {AAAI Press},
url = {https://doi.org/10.1609/aaai.v39i6.32705},
doi = {10.1609/aaai.v39i6.32705},
booktitle = {Proceedings of the Thirty-Ninth AAAI Conference on Artificial Intelligence and Thirty-Seventh Conference on Innovative Applications of Artificial Intelligence and Fifteenth Symposium on Educational Advances in Artificial Intelligence},
articleno = {732},
numpages = {9},
series = {AAAI'25/IAAI'25/EAAI'25}
}

@article{shariff2024spikingdd,
author = {Waseem Shariff  and Paul Kielty  and Joseph Lemley  and Peter Corcoran },
title = {Spiking-DD: neuromorphic event camera based driver distraction detection
               with spiking neural network},
journal = {IET Conference Proceedings},
volume = {2024},
issue = {10},
pages = {71-78},
year = {2024},
doi = {10.1049/icp.2024.3285}
}

@article{zhu2025ultrafastdynamicdefect,
  author = {Zhu, S and Yin, Q and Wang, C and Huang, J and Lam, EY},
  year = {2025},
  month = {November},
  title = {Ultrafast {D}ynamic {D}efect {I}nspection {W}ith {C}omputational {N}euromorphic {I}maging},
  journal = {Adv Sci (Weinh)},
  volume = {12},
  number = {44},
  pages = {e10338},
  doi = {10.1002/advs.202510338},
  pmid = {40985332},
  pmcid = {PMC12667530}
}

@dataset{middleton2026dataset,
  author       = {Middleton, Mike and
                  Trefzer, Martin Albrecht and
                  Kayan, Hakan and
                  Ali, Temoor and
                  Sen Bhattacharya, Basabdatta and
                  Baikas, Epifanios and
                  Charith, Perera and
                  Rhodes, Oliver and
                  Vousden, Mark and
                  Gheorghiu, Elena},
  title        = {ANTShapes Benchmarking Datasets for Event-Based
                   Object Classification
                  },
  month        = aug,
  year         = 2026,
  publisher    = {Zenodo},
  version      = {1.0},
  doi          = {10.5281/zenodo.21813140},
  url          = {https://doi.org/10.5281/zenodo.21813140},
}

\end{document}